\documentclass[10pt,twocolumn,letterpaper]{article}

\usepackage{iccv}
\usepackage{times}
\usepackage{epsfig}
\usepackage{graphicx}
\usepackage{amsmath}
\usepackage{amssymb}
\usepackage{mathtools}
\usepackage{booktabs}
\usepackage{hhline}
\usepackage{multirow}
\usepackage{subfig}
\usepackage[dvipsnames]{xcolor}
\usepackage{url}

\usepackage{pifont}
\newcommand{\cmark}{\ding{51}}%
\newcommand{\xmark}{\ding{55}}%
\newcommand{\good}[1]{{\textcolor{ForestGreen}{#1}}}

\usepackage[breaklinks=true,bookmarks=false]{hyperref}
\usepackage[capitalize]{cleveref}
\crefname{section}{Sec.}{Secs.}
\Crefname{section}{Section}{Sections}
\Crefname{table}{Table}{Tables}
\crefname{table}{Tab.}{Tabs.}
\crefname{equation}{Eq.}{Eqs.}

\newcommand{\minisection}[1]{\noindent{\textbf{#1}}}

\DeclareMathOperator*{\argmax}{arg\,max}
\DeclareMathOperator*{\argmin}{arg\,min}

\iccvfinalcopy 

\def\iccvPaperID{11655} 

\ificcvfinal\fi

\begin{document}

\title{Anti-DreamBooth: Protecting users from personalized text-to-image synthesis}

\author{Thanh Van Le$^{*1}$, Hao Phung$^{*1}$, Thuan Hoang Nguyen$^{*1}$, Quan Dao$^{*1}$, Ngoc N. Tran$^{\dagger2}$, Anh Tran$^1$\\
$^1$VinAI Research\hspace{2cm}$^2$Vanderbilt University\\
{\tt\small v.\{thanhlv19, haopt12, thuannh5, quandm7, anhtt152\}@vinai.io, ngoc.n.tran@vanderbilt.edu}
}


\maketitle
\def\thefootnote{\textsuperscript{$*$}}\footnotetext{Equal contributions.}
\def\thefootnote{\textsuperscript{$\dagger$}}\footnotetext{Work done while at VinAI.}
\ificcvfinal\thispagestyle{empty}\fi

\begin{abstract}
   Text-to-image diffusion models are nothing but a revolution, allowing anyone, even without design skills, to create realistic images from simple text inputs. With powerful personalization tools like DreamBooth, they can generate images of a specific person just by learning from his/her few reference images. However, when misused, such a powerful and convenient tool can produce fake news or disturbing content targeting any individual victim, posing a severe negative social impact. In this paper, we explore a defense system called Anti-DreamBooth against such malicious use of DreamBooth. The system aims to add subtle noise perturbation to each user's image before publishing in order to disrupt the generation quality of any DreamBooth model trained on these perturbed images. We investigate a wide range of algorithms for perturbation optimization and extensively evaluate them on two facial datasets over various text-to-image model versions. Despite the complicated formulation of DreamBooth and Diffusion-based text-to-image models, our methods effectively defend users from the malicious use of those models. Their effectiveness withstands even adverse conditions, such as model or prompt/term mismatching between training and testing. Our code will be available at \href{https://github.com/VinAIResearch/Anti-DreamBooth.git}{https://github.com/VinAIResearch/Anti-DreamBooth.git}. 
\end{abstract}

\section{Introduction}
Within a few years, denoising diffusion models \cite{ho2020denoising, song2020denoising, rombach2022high} have revolutionized image generation studies, allowing producing images with realistic quality and diverse content \cite{dhariwal2021diffusion}. They especially succeed when being combined with language \cite{2020t5} or vision-language models \cite{radford2021learning} for text-to-image generation. Large models \cite{ramesh2022hierarchical,Midjourney,rombach2022high,saharia2022photorealistic,balaji2022ediffi}
can produce photo-realistic or artistic images just from simple text description inputs. A user can now generate art within 
a few seconds, and a generated drawing even beat professional artists in an art competition \cite{Roose2022Sep}. Photo-realistic synthetic images can be hard to distinguish from real photos \cite{Ingram2022Dec}. Besides, ControlNet \cite{zhang2023adding} offers extra options to control the generation outputs, further boosting the power of the text-to-image models and bringing them closer to mass users.

\begin{figure}
    \centering
    \includegraphics[width=.475\textwidth]{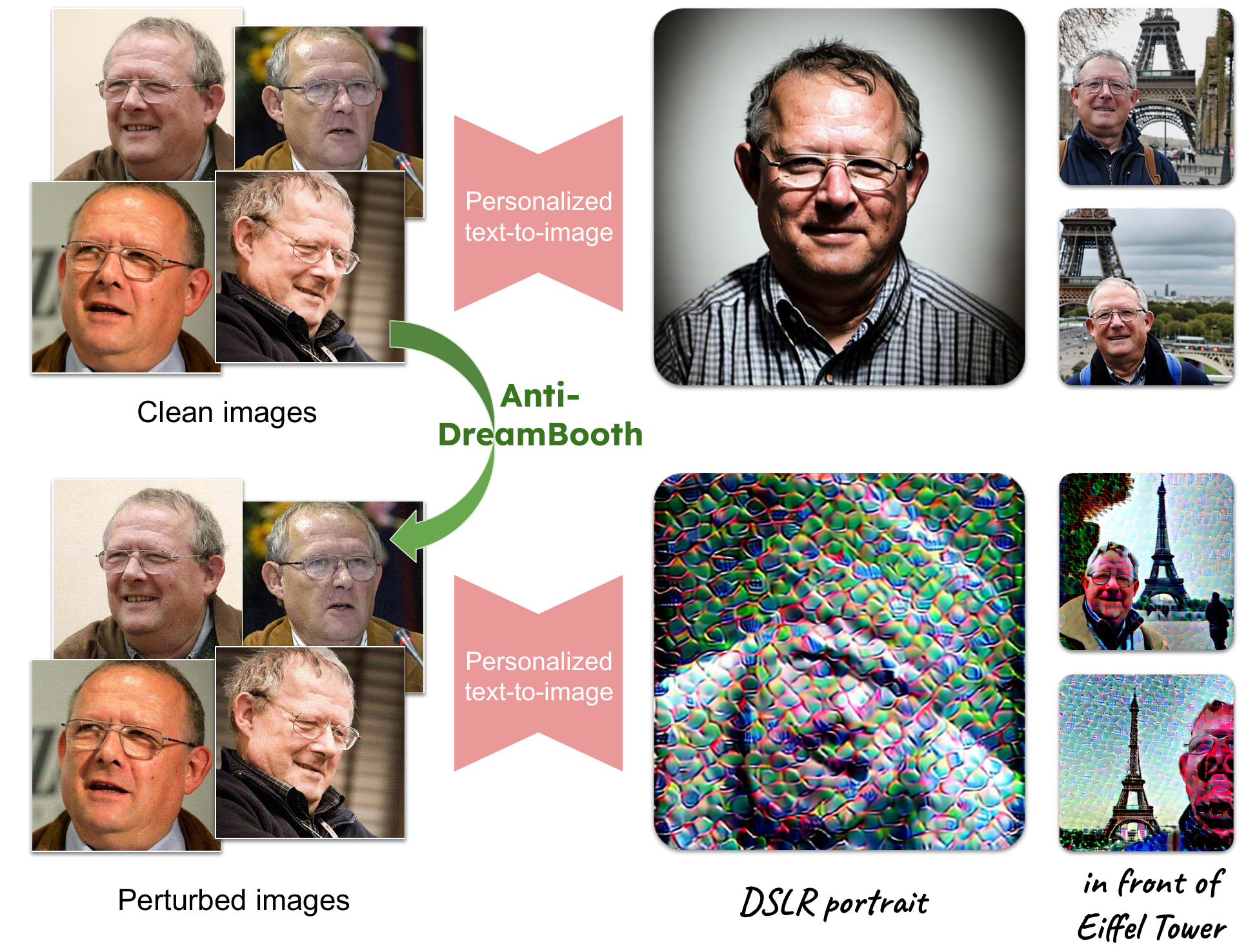}
    \caption{A malicious attacker can collect a user's images to train a personalized text-to-image generator for malicious purposes. Our system, called Anti-DreamBooth, applies imperceptible perturbations to the user's images before releasing, making any personalized generator trained on these images fail to produce usable images, protecting the user from that threat.}
    \label{fig:teaser}
    \vspace{-5mm}
\end{figure}

One extremely useful feature for image generation models is personalization, which allows the models to generate images of a specific subject, given a few reference examples. For instance, one can create images of himself/herself in a fantasy world for fun, or create images of his/her family members as a gift. Textual Inversion \cite{gal2023an} and DreamBooth \cite{ruiz2022dreambooth} are two prominent techniques that offer that impressive ability. While Textual Inversion only optimizes the text embedding inputs representing the target subject, DreamBooth finetunes the text-to-image model itself for better personalization quality. Hence, DreamBooth is particularly popular and has become the core technique in many applications. 

While the mentioned techniques provide a powerful and convenient tool for producing desirable images at will, they also pose a severe risk of being misused. A malicious user can propagate fake news with photo-realistic images of a celebrity generated by DreamBooth. This can be classified as DeepFakes \cite{juefei2022countering}, one of the most serious AI crime threats that has drawn an enormous attention from the media and community in recent years. Besides creating fake news, DreamBooth can be used to issue harmful images targeting specific persons, disrupting their lives and reputations. While the threat of GAN-based DeepFakes techniques is well-known and has drawn much research interest, the danger from DreamBooth has yet to be aware by the community, making its damage, when happening, more dreadful.

This paper discusses how to protect users from malicious personalized text-to-image synthesis. Inspired by DeepFakes's prevention studies \cite{yeh2020disrupting,ruiz2020disrupting,yang2021defending,huang2021initiative,wang2022anti}, we propose to pro-actively defend each user from the DreamBooth threat by injecting subtle adversarial noise into their images before publishing. The noise is designed so that any DreamBooth model trained on these perturbed images fails to produce reasonable-quality images of the target subject. While the proposed mechanism, called Anti-DreamBooth, shares the same goal and objective as the techniques to disrupt GAN-based DeepFakes, it has a different nature due to the complex formulation of diffusion-based text-to-image models and DreamBooth: 
\begin{itemize}
    \item In GAN-based disruption techniques, the defender optimizes the adversarial noise of a single image, targeting a fixed DeepFakes generator. In Anti-DreamBooth, we have to optimize the perturbation noise to disrupt a dynamic, unknown generator that is finetuned from the perturbed images themselves.
    \item GAN-based DeepFakes generator produces each fake image via a single forward step; hence, adversarial noise can be easily learned based on the model's gradient. In contrast, a diffusion-based generator produces each output image via a series of non-deterministic denoising steps, making it impossible to compute the end-to-end gradient for optimization.
    \item Anti-DreamBooth has a more complex setting by considering many distinctive factors, such as the prompt used in training and inference, the text-to-image model structure and pre-trained weights, and more.
\end{itemize}

Despite the complexity mentioned above, we show that the DreamBooth threat can be effectively prevented. Instead of targeting the end-to-end image generation process, we can adapt the adversarial learning process to break each diffusion sampling step. We design different algorithms for adversarial noise generation, and verify their effectiveness in defending DreamBooth attack on two facial benchmarks. Our proposed algorithms successfully break all DreamBooth attempts in the controlled settings, causing the generated images to have prominent visual artifacts. Our proposed defense shows consistent effect when using different text-to-image models and different training text prompts.  More impressively, Anti-DreamBooth maintains its efficiency even under adverse conditions, such as model or prompt/term mismatching between training and testing. 

In summary, our contributions include: (1) We discuss the potential negative impact of personalized text-to-image synthesis, particularly with DreamBooth, and define a new task of defending users from this critical risk, (2) We propose proactively protecting users from the threat by adding adversarial noise to their images before publishing, (3) We design different algorithms for adversarial noise generation, adapting to the step-based diffusion process and finetuning-based DreamBooth procedure, (4) We extensively evaluate our proposed methods on two facial benchmarks and under different configurations. Our best defense works effectively in both convenient and adverse settings.

\section{Related work}
\subsection{Text-to-image generation models} Due to the advent of new large-scale training datasets such as LAION5B \cite{Schuhmann2022LAION5BAO}, text-to-image generative models are advancing rapidly, opening new doors in many visual-based applications and attracting attention from the public. These models can be grouped into four main categories:  auto-regressive \cite{yu2022scaling}, mask-prediction \cite{chang2023muse}, GAN-based \cite{sauer2023stylegan} and diffusion-based approaches, all of which show astounding qualitative and quantitative results. Among these methods, diffusion-based models \cite{rombach2022high, saharia2022photorealistic, balaji2022ediffi, Nichol2021GLIDETP, Ramesh2022HierarchicalTI} have exhibited an exceptional capacity for generating high-quality and easily modifiable images, leading to their widespread adoption in text-to-image synthesis. GLIDE \cite{Nichol2021GLIDETP} is arguably the first to combine a diffusion model with classifier guidance for text-to-image generation. DALL-E 2 \cite{Ramesh2022HierarchicalTI} then improves the quality further using the CLIP text encoder and diffusion-based prior. For better trade-off between efficiency and fidelity, following-up works either introduce coarse-to-fine generation process like Imagen \cite{saharia2022photorealistic} and eDiff-I \cite{balaji2022ediffi} or work on latent space like LDM \cite{rombach2022high}. StableDiffusion \cite{SD}, primarily based on LDM, is the first open-source large model of this type, further boosting the widespread applications of text-to-image synthesis. 


\subsection{Personalization} Customizing the model's outputs for a particular person or object has been a significant aim in the machine-learning community for a long time. Generally, personalized models are commonly observed in recommendation systems \cite{10.1145/3240323.3241729} or federated learning \cite{shamsian2021personalized,SuPerFed}. Within the context of diffusion models, previous research has focused on adapting a pre-trained model to create fresh images based on a particular target idea using natural language cues. Existing methods for personalizing the model either involve adjusting a collection of text embeddings to describe the idea \cite{gal2023an} or fine-tuning the denoising network to connect a less commonly used word-embedding to the novel concept \cite{ruiz2022dreambooth}. For better customization, \cite{kumari2022customdiffusion} propose a novel approach to not only model a new concept of an individual subject by optimizing a small set of parameters of cross-attention layers but also to combine multiple concepts of objects via joint training. Among these tools, DreamBooth \cite{ruiz2022dreambooth} is particularly popular due to its exceptional quality and has become the core technique in many applications. Hence, we focus on defending the malign use of this technique. 


\subsection{Adversarial attacks}
With the introduction of the Fast Gradient Sign Method (FGSM) attack \cite{Goodfellow2014ExplainingAH}, adversarial vulnerability has become an active field of research in machine learning. The goal of adversarial attacks is to generate a model input that can induce a misclassification while remaining visually indistinguishable from a clean one. Following this foundational work, different attacks with different approaches started to emerge, with more notable ones including: \cite{bim, madry2018towards} being FGSM's iterative versions, \cite{cw} limiting the adversarial perturbation's magnitude implicitly using regularization instead of projection, \cite{deepfool} searching for a close-by decision boundary to cross, etc. For black-box attacks, where the adversary does not have full access to the model weights and gradients, \cite{spsa} estimates the gradient using sampling methods, while \cite{boundary, hsja, square} aim to synthesize a close-by example by searching for the classification boundary, then finding the direction to traverse towards a good adversarial example. Combining them, \cite{autoattack} is an ensemble of various attacks that are commonly used as a benchmark metric, being able to break through gradient obfuscation \cite{grad_obf} with the expectation-over-transformation technique \cite{eot}. 


\subsection{User protection with image cloaking}
With the rapid development of AI models, their misuse risk has emerged and become critical. Particularly, many models exploit the public images of each individual for malicious purposes. Instead of passively detecting and mitigating these malign actions, many studies propose proactively preventing them from succeeding. The idea is to add subtle noise into users' images before publishing to disrupt any attempt to exploit those images. This approach is called ``image cloaking'', which our proposed methods belong to.

One application of image cloaking is to prevent privacy violations caused by unauthorized face recognition systems. Fawkes \cite{shan2020fawkes} applies targeted attacks to shift the user's identity towards a different reference person in the embedding space. Although it learns the adversarial noise using a surrogate face recognition model, the noise successfully transfers to break other black-box recognizers. Lowkey \cite{cherepanova2021lowkey} further improves the transferability by using an ensemble of surrogate models. It also considers a Gaussian smoothed version of the perturbed image in optimization, improving robustness against different image transformations. AMT-GAN \cite{hu2022protecting} crafts a natural-looking cloak as makeup, while OPOM \cite{zhong2022opom} optimizes person-specific universal privacy masks.

Another important application of image cloaking is to disrupt GAN-based image manipulation for DeepFakes. Yang et al. \cite{yang2021defending} exploits differentiable image transformations for robust image cloaking. Yeh et al. \cite{yeh2020disrupting} defines new effective objective functions to nullify or distort the image manipulation. Huang et al. \cite{huang2021initiative} addresses personalized DeepFakes techniques by alternating the training of the surrogate model and a perturbation generator. Anti-Forgery \cite{wang2022anti} crafts the perturbation for channels $a$ and $b$ in the $Lab$ color space, aiming for natural-looking and robust cloaking. Lately, UnGANable \cite{li2023unganable} prevents StyleGAN-based image manipulation by breaking its inversion process.

Similar to our goals, two concurrent works, GLAZE \cite{shan2023glaze} and AdvDM \cite{liang2023adversarial}, aim to protect against personalized text-to-image diffusion models exploited without consent using image cloaking. However, our work differs from theirs in several key aspects. First, GLAZE and AdvDM focus on disrupting artistic mimicry, while Anti-DreamBooth concentrates on preventing the negative impacts of generating fake or harmful personalized images. Second, GLAZE utilizes complex style-transferred guidance during noise optimization, which is difficult to adapt to the user protection setting. AdvDM only focuses on a simpler technique Textual Inversion \cite{gal2023an} where the generator is fixed, unlike the challenging finetuning of DreamBooth in our setting.

\section{Problem}
\subsection{Background}
\minisection{Adversarial attacks.}
The goal of adversarial attacks is to find an imperceptible perturbation of an input image to mislead the behavior of given models. Typical works have been developed for classification problems where for a model $f$, an adversarial example $x'$ of an input image $x$ is generated to stay visually undetectable while inducing a misclassification $y_\mathrm{true} \ne f(x')$ (untargeted attack), or making the model predict a predefined target label $y_\text{target} = f(x_\mathrm{adv})\ne y_\mathrm{true}$ (targeted attack). The minimal visual difference is usually enforced by bounding the perturbation to be within an $\eta$-ball w.r.t. an $\ell_p$ metrics, that is $\Vert x' - x\Vert_p<\eta$. To achieve this objective, denoting $\Delta = \{\delta: \Vert \delta \Vert_p \leq \eta \}$, we find the optimal perturbation $\delta$ to maximize the classification loss in the untargeted version:
\begin{equation}
    \delta_{\text{adv}} = \argmax_{\delta \in \Delta} \mathcal{L}(f(x+\delta), y_\text{true}),
\end{equation}
or to minimize the loss for the targeted variant:
\begin{equation}
    \delta_{\text{adv}} = \argmin_{\delta \in \Delta} \mathcal{L}(f(x+\delta), y_\text{target}).
    \label{eq:tar_adv}
\end{equation}





Projected Gradient Descent (PGD) \cite{madry2018towards} is a commonly used attack based on an iterative optimization process. The updating pipeline to predict $x'$ for untargeted attack is:
\begin{align}
\begin{split}
    & x^\prime_0 = x \\
    & x^\prime_{k} = \Pi_{(x,\eta)}(x^\prime_{k-1} + \alpha\cdot\mathrm{sgn}(   \nabla_x\mathcal{L}(f(x^\prime_{k-1}), y_{true})))
\end{split}
\label{eq:pgd}
\end{align}
where $\Pi_{x,\eta}(z)$ restrains pixel values of $z$ within an $\eta$-ball around the original values in $x$. We acquire the adversarial example $x'$ after a pre-defined number of iterations.

\minisection{Diffusion models}
are a type of generative models \cite{sohl2015deep,ho2020denoising} that decouple the role of generation into two opposing procedures: a forward process and a backward process. While the forward process gradually adds noise to an input image until data distribution becomes pure Gaussian noise, the latter learns to reverse that process to obtain the desired data from random noise. Given input image $x_0 \sim q(x)$, the diffusion process perturbs the data distribution with a noise scheduler $\{\beta_t: \beta_t \in (0, 1)\}_{t=1}^T$ producing increasing levels of noise addition through $T$ steps to obtain a sequence of noisy variables: $\{x_1, x_2, \dots, x_T\}$. 
Each variable $x_t$ is constructed via injecting noise at corresponding timestep $t$:
\begin{equation}
    x_{t} = \sqrt{\bar{\alpha_t}}x_0 + \sqrt{1 - \bar{\alpha_t}}\epsilon
\end{equation}
where $\alpha_t = 1 - \beta_t$, $\bar{\alpha_t} = \prod_{s=1}^t \alpha_s$ and $\epsilon \sim \mathcal{N}(0, \mathbf{I})$.


The backward process learns to denoise from noisy variable $x_{t+1}$ to less-noisy variable $x_t$ via simply estimating the injected noise $\epsilon$ with a parametric neural network $\epsilon_\theta(x_{t+1}, t)$.
The denoising process is trained to minimize $\ell_2$ distance between estimated noise and true noise:
\begin{equation}
    \mathcal{L}_{unc}(\theta, x_0) = \mathbb{E}_{x_0, t, \epsilon \in \mathcal{N}(0,1)} \Vert \epsilon - \epsilon_\theta(x_{t+1}, t ) \Vert_2^2
\end{equation}
where $t$ is uniformly samples within $\{1, \dots, T\}$. 



\minisection{Prompt-based Diffusion Models.}
Unlike unconditional diffusion models, prompt-based diffusion models control the sampling process with an additional prompt $c$ to generate photo-realistic outputs which are well-aligned with the text description. The objective is formulated as follows:
\begin{equation}
    \mathcal{L}_{cond}(\theta, x_0) = \mathbb{E}_{x_0, t, c, \epsilon \in \mathcal{N}(0,1)} \Vert \epsilon - \epsilon_\theta(x_{t+1}, t, c) \Vert_2^2
    \label{eq:cond_loss}
\end{equation}
Thanks to prompt condition, the model can produce more excellent performance in terms of  visual quality than its unconditional counterparts. However, the implementation of most prominent methods \cite{saharia2022photorealistic, balaji2022ediffi} are not publicly available. Alternatively, Stable Diffusion has released the pre-trained weights based on Hugging Face implementation \cite{von-platen-etal-2022-diffusers} to facilitate research in the community. Hence, we mainly perform experiments on different versions of Stable Diffusion. 



\minisection{DreamBooth} is a finetuning technique to personalize text-to-image diffusion models for instance of interest. This technique has two aims. First, it enforces the model to learn to reconstruct the user's images, with a generic prompt $c$ such as ``a photo of \textit{sks} [class noun]'', with \textit{sks} is a special term denoting the target user, and ``[class noun]'' is the object type, which can be ``person'' for human subject. To train this, DreamBooth employs the base loss of diffusion models in \cref{eq:cond_loss}, with $x_0$ is each user's reference image. Second, it further introduces a prior preservation loss to prevent overfitting and text-shifting problems when only a small set of instance examples is used. More precisely, it uses a generic prior prompt $c_{pr}$, e.g.,``a photo of [class noun]", and enforces the model to reproduce instance examples randomly generated from that prior prompt using the original weights $\theta_{ori}$. The training loss combines two objectives:
\begin{multline}
    \mathcal{L}_{db}(\theta, x_0) =
    \mathbb{E}_{x_0, t, t'} \Vert \epsilon - \epsilon_\theta(x_{t+1}, t, c) \Vert_2^2 \\ + \lambda \Vert \epsilon' - \epsilon_\theta(x'_{t'+1}, t', c_{pr}) \Vert_2^2
    \label{eq:Ldb}
\end{multline}
where $\epsilon, \epsilon'$ are both sampled from $\mathcal{N}(0, \mathbf{I})$, $x'_{t'+1}$ is noisy variable of class example $x'$ which is generated from original stable diffusion $\theta_{ori}$ with prior prompt $c_{pr}$, and $\lambda$ emphasizes the importance of the prior term. While DreamBooth was originally designed for Imagen~\cite{saharia2022photorealistic}, it was quickly adopted for any text-to-image generator.

\begin{figure*}
    \centering
    \includegraphics[width=.94\textwidth]{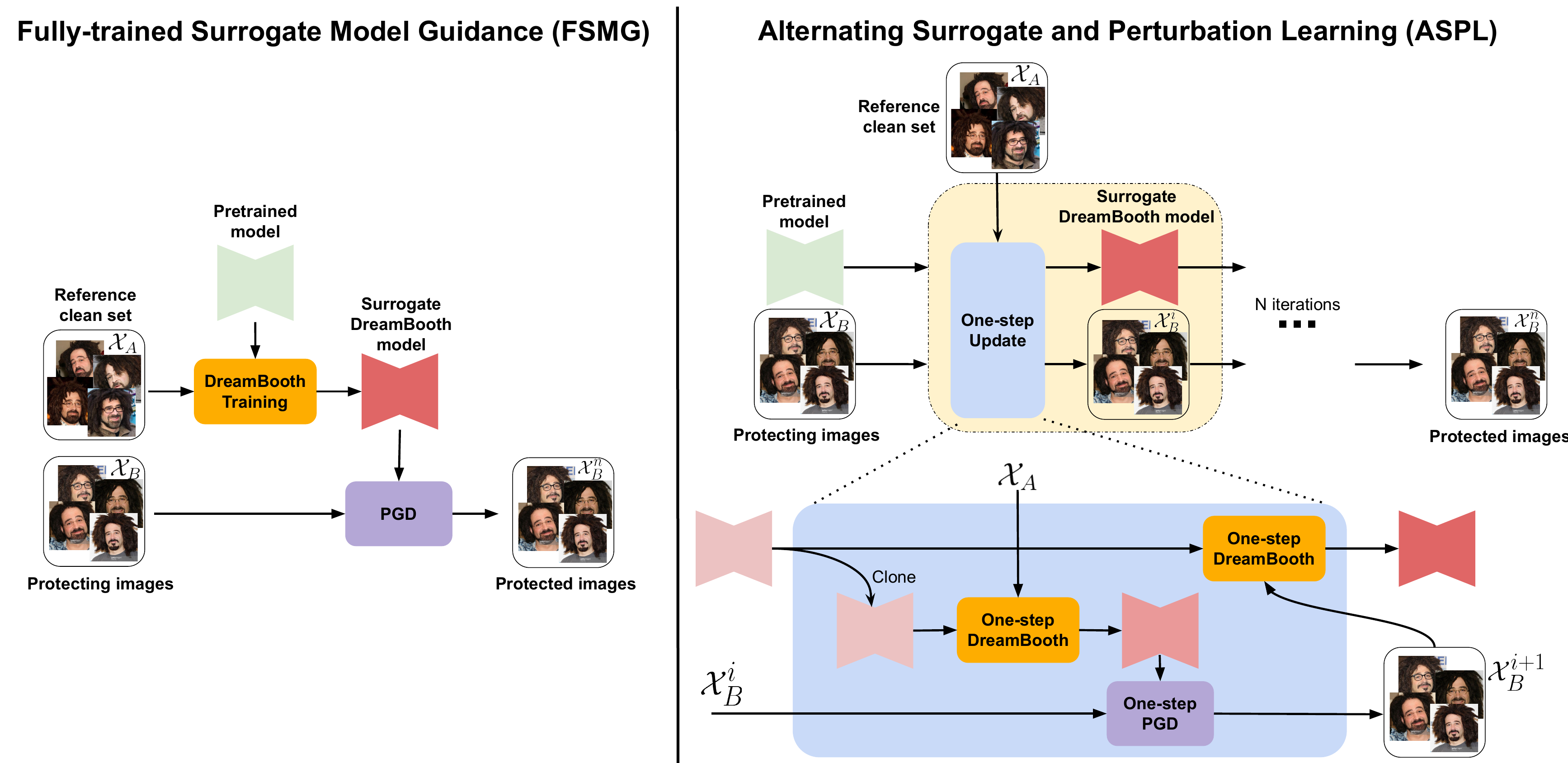}
    \vspace{-2mm}
    \caption{We present here two variants of Anti-DreamBooth, namely Fully-trained Surrogate Model Guidance (FSMG) and Alternating Surrogate and Perturbation Learning (ASPL). Both methods craft the adversarial noise $\delta$ using Projected Gradient Descent (PGD) to maximize the reconstruction loss $\mathcal{L}_{cond}$ of the surrogate model. Left: FSMG uses a fixed surrogate model $\theta_{\text{clean}}$ fully finetuned on a small clean image set $\mathcal{X}_A$ to guide the PGD optimization. Right: ASPL alternates between (i) finetuning a clone surrogate model $\theta'$ on clean images $\mathcal{X}_A$, and (ii) using this clone model to craft $\delta$ for the current image set $\mathcal{X}_B^i$ via PGD. The actual surrogate model $\theta$ is then finetuned on the perturbed images $\mathcal{X}_B^{i+1}$ before the next iteration.}
    \label{fig:system_figure}
    \vspace{-5mm}
\end{figure*}

\subsection{Problem definition}
As a powerful tool to generate photo-realistic outputs of a target instance, DreamBooth can be a double-edged sword. When misused, it can generate harmful images toward the target individual. To mitigate this phenomenon, we propose to craft an imperceptible perturbation added to each user's image that can disrupt the finetuned DreamBooth models to generate distorted images with noticeable artifacts. We define the problem formally below.

Denote $\mathcal{X}$ as the set of images of the person to protect. For each image $x \in \mathcal{X}$, we add an adversarial perturbation $\delta$ and publish the modified image $x' = x + \delta$, while keeping the original one private. The published image set is called $\mathcal{X}'$. An adversary can collect a small image set of that person $\mathcal{X}'_{db} = \{x^{(i)} + \delta^{(i)} \}_{i=1}^{N_{db}} \subset \mathcal{X}'$. He then uses that set as a reference to finetune a text-to-image generator $\epsilon_{\theta}$, following the DreamBooth algorithm, to get the optimal hyper-parameters $\theta^*$.
The general objective is to optimize the adversarial noise $\Delta_{db} = \{\delta^{(i)}\}_{i=1}^{N_{db}}$ that minimizes the personalized generation ability of that DreamBooth model:
\begin{align}
\begin{split}
    & \Delta^*_{db}  = \argmin_{\Delta_{db}} {\mathcal{A}(\epsilon_{\theta^*}, \mathcal{X}}),\\
    \text{s.t.} \quad & \theta^*  = \argmin_{\theta} \sum_{i=1}^{N_{db}} \mathcal{L}_{db}(\theta, x^{(i)} + \delta^{(i)}),\\
    \text{and} \quad & \Vert \delta^{(i)} \Vert_p \leq \eta \quad \forall i \in \{1, 2,..,N_{db}\},
\end{split}
\label{eq:objective}
\end{align}
where $\mathcal{L}_{db}$ is defined in Eq. \ref{eq:Ldb} and $\mathcal{A}(\epsilon_{\theta^*}, \mathcal{X})$ is some personalization evaluation function that assesses the quality of images generated by the DreamBooth model $\epsilon_{\theta^*}$ and the identity correctness based on the reference image set $\mathcal{X}$.

However, it is hard to define a unified evaluation function $\mathcal{A}$. A defense succeeds when the DreamBooth-generated images satisfy one of the criteria: (1) awful quality due to extreme noise, blur, distortion, or noticeable artifacts, (2) none or unrecognizable human subjects, (3) mismatched subject identity. Even with the first criteria, there is no all-in-one image quality assessment metric. Instead, we can use simpler objective functions disrupting the DreamBooth training to achieve the same goal.

We further divide the defense settings into categories, from easy to hard: convenient, adverse, and uncontrolled.

\minisection{Convenient setting.} In this setting, we have prior knowledge about the pretrained text-to-image generator, training term (e.g., ``sks''), and training prompt $c$ the attacker will use. While sounding restricted, it is practical. First, the pretrained generator has to be high-quality and open-source. 
So far, only Stable Diffusion has been made publicly available with several versions released. Second, people often use the default training term and prompt provided in DreamBooth's code. This setting is considered as ``white-box''. 

\minisection{Adverse settings.} In these settings, the pretrained text-to-image generator, training term, or training prompt used by the adversary is unknown. The defense method, if needed, can use a surrogate component that potentially mismatches the actual one to craft the adversarial noises. These settings can be considered as ``gray-box''.

\minisection{Uncontrolled setting.} This is an extra, advanced setting in which some of the user's clean images are leaked to the public without our control. The adversary, therefore, can collect a mix of perturbed and clean images $\mathcal{X}'_{db} = \mathcal{X}'_{adv} \cup \mathcal{X}_{cl}$, with $\mathcal{X}'_{adv} \subset \mathcal{X}'$ and $\mathcal{X}_{cl} \subset \mathcal{X}$. This setting is pretty challenging since the DreamBooth model can learn from unperturbed photos to generate reasonable personalized images.


\section{Proposed defense methods}
\subsection{Overall direction}
As discussed, 
we can aim to attack the learning process of DreamBooth. As the DreamBooth model overfits the adversarial images, we can trick it into performing worse in reconstructing clean images:
\begin{align}
\begin{split}
    \delta^{*(i)} = & \argmax_{\delta^{(i)}} \mathcal{L}_{cond}(\theta^*, x^{(i)}), \forall i \in \{1,..,N_{db}\},\\
    \text{s.t.} \quad & \theta^*  = \argmin_{\theta} \sum_{i=1}^{N_{db}} \mathcal{L}_{db}(\theta, x^{(i)} + \delta^{(i)}),\\
    \text{and} \quad & \Vert \delta^{(i)} \Vert_p \leq \eta \quad \forall i \in \{1,..,N_{db}\},
\end{split}
\label{eq:objective_new}
\end{align}
where $\mathcal{L}_{cond}$ and $\mathcal{L}_{db}$ are defined in Eq. \ref{eq:cond_loss} and \ref{eq:Ldb}. Note that, unlike traditional adversarial attacks, our loss functions are computed only at a randomly-chosen timestep in the denoising sequence during training. Still, this scheme is effective in breaking the generation output (Sec. \ref{sec:exp}). 

\subsection{Algorithms}\label{sec:algorithms}
The problem in Eq. \ref{eq:objective_new} is still a challenging bi-level optimization. We define different methods to approximate its solution based on prominent techniques used in literature.

\minisection{Fully-trained Surrogate Model Guidance (FSMG).}
\label{minisec:fsmg}
Most previous image cloaking approaches \cite{shan2020fawkes,shan2020fawkes,yeh2020disrupting} employ a model trained on clean data as a surrogate to guide the adversarial attack. We can naively follow that direction by using a surrogate DreamBooth model with hyper-parameters $\theta_{\text{clean}}$ fully finetuned from a small subset of samples $\mathcal{X}_A \subset \mathcal{X}$. This set does not need to cover the target images $\mathcal{X}_{db} = \{x^{(i)}\}_{i=1}^{N_{db}}$; it can be fixed, allowing the surrogate model to be learned once regardless of the constant change of $\mathcal{X}$ and $\mathcal{X}_{db}$. After getting $\theta_{clean}$, we can use it as guidance to find optimal noise for each target image $\delta^{*(i)} = \argmax_{\delta^{(i)}} \mathcal{L}_{cond}(\theta_{\text{clean}}, x^{(i)} + \delta^{(i)})$. By doing so, we expect any DreamBooth model finetuned from the perturbed samples to stay away from $\theta_{clean}$.

\minisection{Alternating Surrogate and Perturbation Learning (ASPL).}
Using a surrogate model full-trained on clean data may not be a good approximation to solve the problem in Eq. \ref{eq:objective_new}. Inspired by \cite{huang2021initiative}, we propose to incorporate the training of the surrogate DreamBooth model with the perturbation learning in an alternating manner. The surrogate model $\epsilon_\theta$ is first initiated with the pretrained weights. In each iteration, a clone version $\epsilon'_{\theta'}$ is finetuned on the reference clean data $\mathcal{X}_{A}$, following \cref{eq:Ldb}. This model is then utilized to expedite the learning of adversarial noises $\delta^{(i)}$ in the current loop. Finally, we update the actual surrogate model $\epsilon_\theta$ on the updated adversarial samples, and move to the next training iteration. We provide a snippet for one training iteration in Eq. \ref{eq:ASPL}. With such a procedure, the surrogate model better mimics the true models trained by the malicious DreamBooth users since it is only trained on perturbed data.
\begin{align}
\begin{split}
    \theta' \  \  \: & \leftarrow \theta.\text{clone()}\\
    \theta' \  \  \: & \leftarrow \argmin_{\theta'} \sum_{x \in \mathcal{X}_A} \mathcal{L}_{db}(\theta', x)\\
    \delta^{(i)} & \leftarrow \argmax_{\delta^{(i)}} \mathcal{L}_{cond}(\theta', x^{(i)}+\delta^{(i)})\\ 
    \theta \  \  \: & \leftarrow \argmin_{\theta} \sum_{i=1}^{N_{db}} \mathcal{L}_{db}(\theta, x^{(i)} + \delta^{(i)}).
\end{split}
\label{eq:ASPL}
\end{align}
\begin{table*}[t]
    \centering
    \setlength{\tabcolsep}{4pt}
    \begin{tabular}{l|l|cccc|cccc}
        \toprule
        \multirow{2}{*}{Dataset} & \multirow{2}{*}{Method} & \multicolumn{4}{c|}{``a photo of \textit{sks} person''} & \multicolumn{4}{c}{``a dslr portrait of \textit{sks} person''}\\
        \cline{3-10}
         & & FDFR$\uparrow$ & ISM$\downarrow$ & SER-FQA$\downarrow$ & BRISQUE$\uparrow$ & FDFR$\uparrow$ & ISM$\downarrow$ & SER-FQA$\downarrow$ & BRISQUE$\uparrow$ \\
        \hline
        \multirow{5}{*}{VGGFace2} & No Defense &0.07 &0.63 &0.73 &15.61 &0.21 &0.48 &0.71 &9.64 \\
         & FSMG &0.56 &\textbf{0.33} &\textbf{0.31} &\textbf{36.61} &0.62 &0.29 &0.37 &38.22 \\
         & ASPL &\textbf{0.63} &\textbf{0.33} &\textbf{0.31} &36.42 &\textbf{0.76} &\textbf{0.28} &\textbf{0.30} &\textbf{39.00} \\
         & T-FSMG &0.07 &0.58 &0.74 &15.49 &0.28 &0.44 &0.71 &17.29 \\
         & T-ASPL &0.07 &0.57 &0.72 &15.36 &0.39 &0.44 &0.70 &20.06 \\
        \hline
        \multirow{5}{*}{CelebA-HQ} & No Defense &0.10 &0.68 &0.72 &17.06 &0.26 &0.44 &0.72 &7.30 \\
         & FSMG &\textbf{0.34} &\textbf{0.48} &0.56 &36.13 &\textbf{0.35} &\textbf{0.36} &0.66 &33.60 \\
         & ASPL &0.31 &0.50 &\textbf{0.55} &\textbf{38.57} &0.34 &0.39 &\textbf{0.63} &\textbf{34.89} \\
         & T-FSMG &0.06 &0.64 &0.73 &25.75 &0.24 &0.45 &0.73 &8.04 \\
         & T-ASPL &0.06 &0.64 &0.73 &20.58 &0.26 &0.46 &0.72 &5.36 \\
        \bottomrule
    \end{tabular}
    \vspace{-2mm}
    \caption{Comparing the defense performance of the proposed methods in a convenient setting on different datasets.}
    \label{tab:convenient}
    \vspace{-5mm}
\end{table*}

\begin{figure*}[t] 
    \centering
    \subfloat[Comparison between proposed methods]{
        \includegraphics[height=5.5cm]{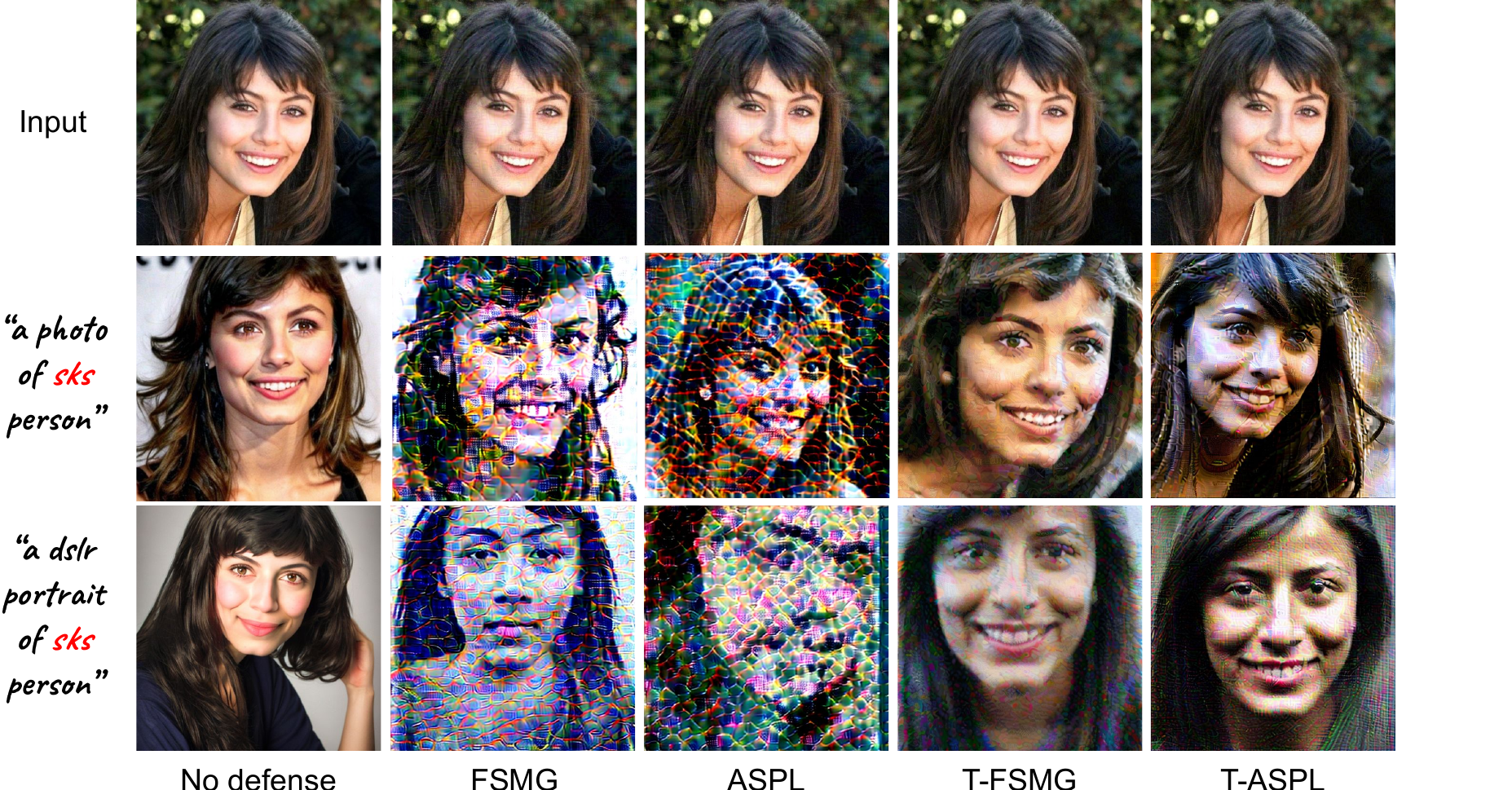}
        \label{fig:compare}
        \vspace{-3mm}
    }
    \hspace{-0.3cm}
    \subfloat[Inference prompts]{
        \includegraphics[height=5.5cm]{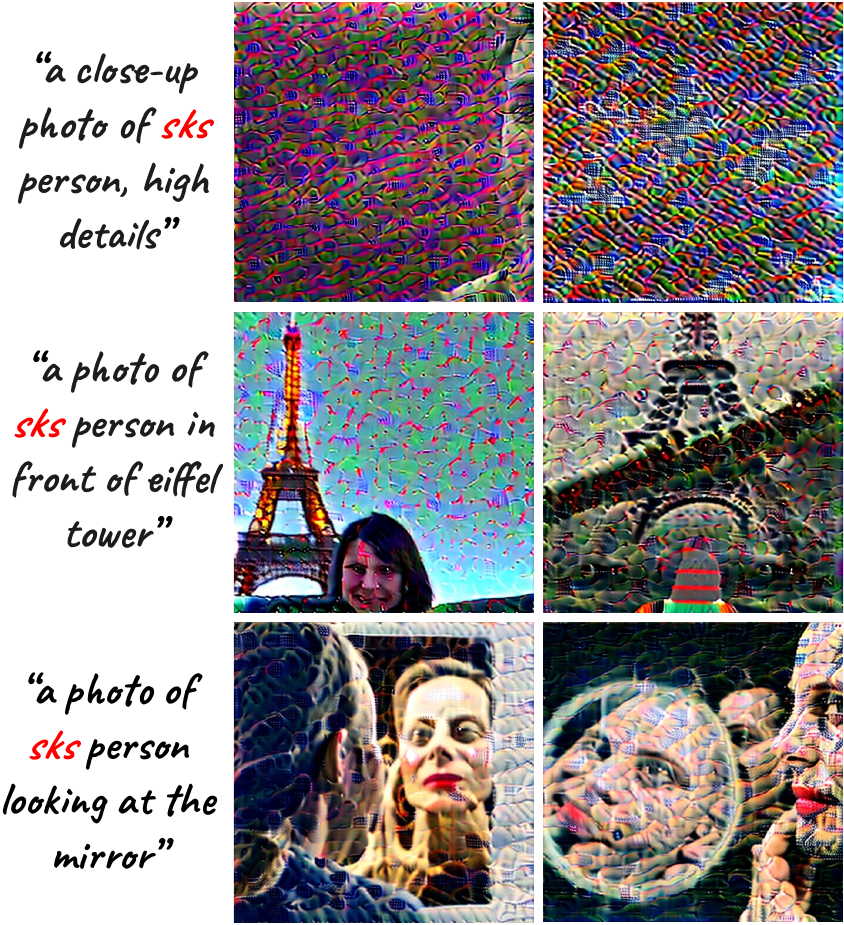}
        \label{fig:prompts}
        \vspace{-3mm}
    }
        \hspace{0.1cm}
        \vspace{-2mm}
    \caption{Qualitative defense results for two subjects in VGGFace2 in the convenient setting. Best viewed in zoom.}
        \vspace{-4mm}
    \label{fig:qual}
\end{figure*}

\begin{table*}[t]
    \centering
    \setlength{\tabcolsep}{3.5pt}
    \begin{tabular}{l|c|cccc|cccc}
        \specialrule{.09em}{.04em}{.04em} 
         \multirow{2}{*}{Version} & \multirow{2}{*}{Defense?} & \multicolumn{4}{|c|}{``a photo of \textit{sks} person''} & \multicolumn{4}{c}{``a dslr portrait of \textit{sks} person''}\\
         \cline{3-10}
          &  & FDFR$\uparrow$ & ISM$\downarrow$ & SER-FQA$\downarrow$ & BRISQUE$\uparrow$ & FDFR$\uparrow$ & ISM$\downarrow$ & SER-FQA$\downarrow$ & BRISQUE$\uparrow$ \\
         \hline
         \multirow{2}{*}{v1.4} & \xmark &0.05 &0.46 &0.65 &21.06 &0.08 &0.43 &0.64 &10.05 \\
          & \cmark &\textbf{0.80} &\textbf{0.18} &\textbf{0.12} &\textbf{26.76} &\textbf{0.17} &\textbf{0.28} &\textbf{0.55} &\textbf{13.07}  \\
          \hline
         \multirow{2}{*}{v1.5} & \xmark &0.07 &0.49 &0.65 &18.53 &0.07 &0.45 &0.64 &10.57 \\
          & \cmark &\textbf{0.71} &\textbf{0.20} &\textbf{0.20} &\textbf{22.98} &\textbf{0.11} &\textbf{0.26} &\textbf{0.57} &\textbf{16.10}  \\
        \specialrule{.09em}{.04em}{.04em} 
    \end{tabular}
    \vspace{-3mm}
    \caption{Defense performance of ASPL with different generator versions on VGGFace2 in a convenient setting.}
    \label{tab:versions}
    \vspace{-1mm}
\end{table*}

\begin{table*}[t]
    \centering
    \setlength{\tabcolsep}{3pt}
    \begin{tabular}{c|cc|cccc|cccc}
        \hline
         \multirow{2}{*}{$\eta$} & \multicolumn{2}{|c|}{Quality} & \multicolumn{4}{|c|}{``a photo of \textit{sks} person''} & \multicolumn{4}{c}{``a dslr portrait of \textit{sks} person''}\\
         \cline{2-9}
          & PSNR$\uparrow$ & LPIPS$\downarrow$ & FDFR$\uparrow$ & ISM$\downarrow$ & SER-FQA$\downarrow$ & BRISQUE$\uparrow$ & FDFR$\uparrow$ & ISM$\downarrow$ & SER-FQA$\downarrow$ & BRISQUE$\uparrow$ \\
         \hline
         0 & - & - & 0.07 &0.63 &0.73 &15.61 &0.21 &0.48 &0.71 &9.64 \\
         0.01 & 48.74 & 0.01 & 0.08 &0.58 &0.72 &33.03 &0.28 &0.45 &0.72 &17.14 \\
         0.03 & 38.42 & 0.12 & 0.44 &0.38 &0.38 &36.45 &0.55 &0.32 &0.43 &37.86 \\
         $0.05^*$ & 34.56 & 0.21 & 0.63 &0.33 &0.31 &36.42 &0.76 &0.28 &0.30 &39.00  \\
         0.10 & 28.77 & 0.40 & 0.76 &0.21 &0.22 &37.33 &0.86 &0.23 &0.26 &40.92 \\
         0.15 & 25.97 & 0.50 & \textbf{0.80} &\textbf{0.15} &\textbf{0.15} &\textbf{37.07} &\textbf{0.91} &\textbf{0.17} &\textbf{0.14} &\textbf{41.18} \\
         \hline
    \end{tabular}
    \vspace{-3mm}
    \caption{Quality of protected images and defense performance of ASPL with different noise budgets on VGGFace2 in a convenient setting. ``*'' is default.}
    \vspace{-3mm}
    \label{tab:noise}
\end{table*}

\begin{table*}[t]
    \centering
    \setlength{\tabcolsep}{2.5pt}
    \begin{tabular}{l|c|c|cccc|cccc}
        \hline
          & \multirow{2}{*}{Train} & \multirow{2}{*}{Test} & \multicolumn{4}{|c|}{``a photo of \textit{sks} person''} & \multicolumn{4}{c}{``a dslr portrait of \textit{sks} person''}\\
         \cline{4-11}
          & & & FDFR$\uparrow$ & ISM$\downarrow$ & SER-FQA$\downarrow$ & BRISQUE$\uparrow$ & FDFR$\uparrow$ & ISM$\downarrow$ & SER-FQA$\downarrow$ & BRISQUE$\uparrow$ \\
         \hline
         \multirow{2}{.076\linewidth}{Model mismatch} & v1.4 & v2.1 &0.62 &0.31 &\textbf{0.28} &36.00 &0.70 &0.31 &0.35 &38.39 \\
         & v1.4 & v2.0 &0.70 &0.27 &0.23 &36.83 &0.61 &0.26 &0.31  &37.28 \\
         \hline
         \multirow{2}{.076\linewidth}{Ensemble} & v1.4, 1.5, 2.1 & v2.0 & \textbf{0.79} & \textbf{0.24} & \textbf{0.18} &\textbf{37.96} &\textbf{0.71} &\textbf{0.23} &\textbf{0.23} &\textbf{38.99} \\

         & v1.4, 1.5, 2.1 & v2.1 &\textbf{0.70} & \textbf{0.27} &\textbf{0.28} &\textbf{36.71} &\textbf{0.75} &\textbf{0.29} &\textbf{0.33} &\textbf{39.23} \\

         \hhline{===========}
         \multirow{4}{.076\linewidth}{Term/ Prompt mismatch} & \multicolumn{2}{|c|}{\multirow{2}{*}{DreamBooth prompt}} & \multicolumn{4}{|c|}{``a photo of \textit{$S_*$} person''} & \multicolumn{4}{c}{``a dslr portrait of \textit{$S_*$} person''}\\
         \cline{4-11}
          & \multicolumn{2}{|c|}{} &FDFR$\uparrow$ & ISM$\downarrow$ & SER-FQA$\downarrow$ & BRISQUE$\uparrow$ & FDFR$\uparrow$ & ISM$\downarrow$ & SER-FQA$\downarrow$ & BRISQUE$\uparrow$ \\
         \cline{2-11}
         & \multicolumn{2}{|c|}{``\textit{sks}'' $\rightarrow$ ``\textit{t@t}''} &0.34 &0.30 &0.48 &36.67 &0.34 &0.28 &0.52  &28.17 \\
         & \multicolumn{2}{|c|}{``a dslr portrait of \textit{sks} person''} &0.07 &0.15 &0.69 &17.34 &0.49 &0.37 &0.36 &38.42 \\
         \hline
    \end{tabular}
    \vspace{-3mm}
    \caption{Defense performance of ASPL on VGGFace2 when the model, term, or prompt used to train the target DreamBooth model is different from the one used to generate defense noise and when the ensemble technique is applied. Here, \textit{$S_*$} is ``t@t'' for the first row and ``sks'' for second row.}
    \label{tab:cross}
    \vspace{-1mm}
\end{table*}

\begin{table*}[t]
    \centering
    \setlength{\tabcolsep}{3pt}
    \begin{tabular}{c|c|cccc|cccc}
        \hline
         \multirow{2}{*}{Perturbed} & \multirow{2}{*}{Clean} & \multicolumn{4}{|c|}{``a photo of \textit{sks} person''} & \multicolumn{4}{c}{``a dslr portrait of \textit{sks} person''}\\
         \cline{3-10}
          & & FDFR$\uparrow$ & ISM$\downarrow$ & SER-FQA$\downarrow$ & BRISQUE$\uparrow$ & FDFR$\uparrow$ & ISM$\downarrow$ & SER-FQA$\downarrow$ & BRISQUE$\uparrow$ \\
         \hline
         4&0 &\textbf{0.63} &\textbf{0.33} &\textbf{0.31} &\textbf{36.42} &\textbf{0.76} &\textbf{0.28} &\textbf{0.30} &\textbf{39.00} \\
         \hline
         3&1 &0.50 &0.43 &0.41 &35.53 &0.52 &0.35 &0.51 &34.01 \\
         2&2 &0.29 &0.53 &0.61 &28.99 &0.40 &0.37 &0.62 &26.13 \\
         1&3 &0.08 &0.61 &0.73 &18.92 &0.27 &0.45 &0.70 &15.55 \\
         \hline
         0&4 &0.07 &0.63 &0.73 &15.61 &0.21 &0.48 &0.71 &9.64\\
         \hline
    \end{tabular}
    \vspace{-3mm}
    \caption{Defense performance of ASPL on VGGFace2 in uncontrolled settings. We include two extra results with 0 clean image (convenient setting) and 0 perturbed image (no defense) for comparison. } 
    \vspace{-3mm}
    \label{tab:uncontrolled}
\end{table*}

\begin{table}[]
\resizebox{\columnwidth}{!}{
\begin{tabular}{|l|l|l|l|l|}
\hline
 &
  \begin{tabular}[c]{@{}l@{}}FDFR$ \uparrow$\end{tabular} &
  \begin{tabular}[c]{@{}l@{}}ISM$\downarrow$\end{tabular} &
  \begin{tabular}[c]{@{}l@{}}SER-FQA$\downarrow$\end{tabular} &
  \begin{tabular}[c]{@{}l@{}}BRISQUE$\uparrow$\end{tabular} \\ \hline
ASPL                 & \textbf{0.63} & \textbf{0.33} & \textbf{0.31} & \textbf{36.42} \\ \hline
Gaussian Blur K=3    & \textbf{0.48} & \textbf{0.42} & \textbf{0.39} & \textbf{42.05} \\
Gaussian Blur K=5    & 0.19 & 0.51 & 0.62 & \textbf{42.46} \\
Gaussian Blur K=7    & 0.10 & 0.56 & 0.68 & \textbf{43.72} \\
Gaussian Blur K=9    & 0.07 & 0.59 & 0.71 & \textbf{40.67} \\ \hline
JPEG Comp. Q=10      & 0.09 & 0.58 & 0.71 & \textbf{43.93} \\
JPEG Comp. Q=30      & 0.08 & 0.59 & 0.73 & \textbf{32.56} \\
JPEG Comp. Q=50      & 0.11 & 0.56 & 0.70 & \textbf{30.29} \\
JPEG Comp. Q=70      & 0.19 & 0.49 & 0.56 & \textbf{37.04} \\ \hline
No def., no preproc. & 0.07 & 0.63 & 0.73 & 15.61 \\ \hline
\end{tabular}
}
\vspace{-3mm}
\caption{ASPL’s performance on VGGFace2, using the prompt ``\textit{A photo of sks person}''.} 
\label{tab:robustness_preprocessing}
\end{table}

\begin{table}[]
\resizebox{\columnwidth}{!}{%
\begin{tabular}{|l|l|l|l|l|l|}
\hline
     & Def.?  & FDFR$\uparrow$ & ISM$\downarrow$ & SER-FQA$\downarrow$ & BRISQUE$\uparrow$ \\ \hline
TI   & \xmark & 0.06           & 0.50            & 0.67                & 7.79              \\
TI   & \cmark & \textbf{0.43}           & \textbf{0.12}            & 0.59                & \textbf{36.79}             \\ \hline
LoRA & \xmark & 0.06           & 0.52            & 0.69                & 17.25             \\
LoRA & \cmark & \textbf{0.64}           & \textbf{0.23}            & \textbf{0.27}                & \textbf{42.07}             \\ \hline
\end{tabular}%
}
\vspace{-3mm}
\caption{ASPL's performance against Textual Inversion and LoRA DreamBooth, the prompt is ``\textit{A photo of sks person}''.}
\vspace{-3mm}
\label{tab:ti_lora}
\end{table}


\minisection{Targeted approaches.}
The proposed algorithms above are untargeted; each perturbation noise is learned to maximize the reconstruction loss $\mathcal{L}_{cond}$. Therefore, the adversarial examples $x^{(i)} + \delta^{(i)}$ may guide the target DreamBooth model to learn different adversarial directions, potentially canceling out their effects. Inspired by the success of targeted attacks in \cite{shan2020fawkes}, we can select a single target $x^{tar}$ to learn optimal $\delta$ such that the output of model is pulled closer to $x^{tar}_t$ when trained with $(x+\delta)_t$. This targeted attack scheme can be plugged into all previous methods, and we denote new algorithms with the prefix ``T-'', e.g., T-FSMG and T-ASPL.


\minisection{Ensemble approaches.}
In adverse settings, the pretrained text-to-image generator used by the attacker is unknown. While we can pick one to train the perturbation and hope it transfers well to the target generator, one better approach is to use an ensemble \cite{cherepanova2021lowkey,yang2021defending} of surrogate models finetuned from different pretrained generators. This approach can be an easy plug-in for the previous approaches. Due to memory constraints, instead of using these surrogate models all at once, we only used a single model at a time, in an interleaving manner, to produce optimal perturbed data.

\section{Experiments}\label{sec:exp}
\subsection{Experimental setup}\label{sec:Setup}
\minisection{Datasets.}
To evaluate the effectiveness of the proposed methods, we look for facial benchmarks that satisfy the following criteria: (1) each dataset covers a large number of different subjects with identity annotated, (2) each subject must have enough images to form two image sets for reference ($\mathcal{X}_A$) and protection ($\mathcal{X}_{db}$), (3) the images should have mid- to high-resolution, (4) the images should be diverse and in-the-wild. Based on those criteria, we select two famous face datasets CelebA-HQ \cite{karras2017progressive} and VGGFace2 \cite{Cao18}. 

CelebA-HQ is a high-quality version of CelebA \cite{liu2015faceattributes} that consists of $30,000$ images at $1024 \times 1024$ resolution. We use the annotated subset \cite{celeba_id} that filters and groups images into $307$ subjects with at least $15$ images for each subject.

VGGFace2 \cite{Cao18} contains around $3.31$ million images of $9131$ person identities. We filter the dataset to pick subjects that have at least $15$ images of resolution above $500\times500$. 

For fast but comprehensive evaluations, we choose $50$ identities for each dataset. For each subject in these datasets, we use the first $12$ images and divide them into three subsets, including the reference clean image set, the target protecting set, and an extra clean image set for uncontrolled setting experiments (Sec. \ref{sec:uncontrolled}). Each mentioned subset has $4$ images with diverse conditions. We then center-crop and resize images to resolution $512 \times 512$.

\minisection{Training configurations.}
We train each DreamBooth model, both text-encoder and UNet model, with batch size of $2$ and learning rate of $5\times10^{-7}$ for $1000$ training steps. By default, we use the latest Stable Diffusion (v2.1) as the pretrained generator. Unless specified otherwise, the training instance prompt and prior prompt are ``a photo of \textit{sks} person'' and ``a photo of person'', respectively. It takes 15 minutes to train a model on an NVIDIA A100 GPU 40GB.

We optimize the adversarial noise $\delta^{(i)}$ in each step of FSMG and ASPL using the untargeted PGD scheme (Eq. \ref{eq:pgd}). 
We use 100 PGD iterations for FSMG and 50 iterations for ASPL.
Both methods use $\alpha = 0.005$ and the default noise budget $\eta = 0.05$. It takes 2 and 5 minutes for FSMG and ASPL to complete on an NVIDIA A100 GPU 40GB.

\minisection{Evaluation metrics.}
Our methods aim to disrupt the target DreamBooth models, making them produce poor images of the target user. To measure the defense's effectiveness, for each trained DreamBooth model and each testing prompt, we generate 30 images. We then use a series of metrics to evaluate these generated images comprehensively. 

Images generated from successfully disrupted DreamBooth models may have no detectable face, and we measure that rate, called \textbf{\textit{Face Detection Failure Rate (FDFR)}}, using RetinaFace detector \cite{deng2020retinaface}. If a face is detected, we extract its face recognition embedding, using ArcFace recognizer \cite{deng2019arcface}, and compute its cosine distance to the average face embedding of the entire user's clean image set. This metric is called \textbf{\textit{Identity Score Matching (ISM)}}. Finally, we use two extra image quality assessment metrics. One is \textbf{\textit{SER-FQA}} \cite{TerhorstKDKK20}, which is an advanced, recent metric dedicated to facial images. The other is \textbf{\textit{BRISQUE}} \cite{Brisque}, which is classical and popular for assessing images in general.

\subsection{Convenient setting}
We first evaluate the proposed defense methods, including FSMG, ASPL, T-FSMG, and T-ASPL, in a convenient setting on the two datasets. We try two image generation prompts, one used in training (``a photo of \textit{sks} person'') and one novel, unseen prompt (``a dslr portrait of \textit{sks} person''). The average scores over DreamBooth-generated images with each defense are reported in Table \ref{tab:convenient}. As can be seen, the untargeted defenses significantly increase the face detection failure rates and decrease the identity matching scores, implying their success in countering the DreamBooth threat. We provide some qualitative images in Fig. \ref{fig:compare}. As expected, ASPL defends better than FSMG since it mimics better the DreamBooth model training at test time. Targeted methods perform poorly, suggesting that the noise generated by these methods, while providing more consistent adversarial guidance in DreamBooth training, is suboptimal and  ineffective. Since ASPL performs the best, we will try only this method in all follow-up experiments.

\minisection{Qualitative Evaluation.} We further evaluate ASPL via a user study with 40 participants. For each of 50 identities in VGGFace2, we train two DreamBooth models on (1) original and (2) ASPL-perturbed images. From each model, we generate 6 images with the prompt ``\textit{A photo of sks person}''. Users are asked to input the numbers of good-quality and correct-identity images for each set. The average numbers are used as evaluation scores (lower is better). ASPL significantly degraded the quality (4.72 to 0.42) and identity preservation (3.70 to 1.16) of the generated images, confirming our method's effectiveness.

\subsection{Ablation studies}

\minisection{Text-to-image generator version.} In previous experiments, we used Stable Diffusion (SD) v2.1 as the pretrained text-to-image generator. In this section, we examine if our proposed defense (ASPL) is still effective when using different pretrained generators. Since SD is the only open-source large text-to-image model series, we try two of its versions, including v1.4 and v1.5. Note that while belonging to the same model family, these models, including v2.1, are slightly different in networks and behaviors. As reported in Table \ref{tab:versions}, ASPL shows consistent defense effectiveness.

\minisection{Noise budget.} Next, we examine the impact of noise budget $\eta$ on ASPL attack using SD v2.1. As illustrated in Table \ref{tab:noise}, our defense is already effective with $\eta = 0.03$. The larger the noise budget is, the better defense performance we get, at the cost of the perturbation's stealthiness. 

\minisection{Inference text prompt.} As indicated in Table \ref{tab:convenient}, ASPL well-disturbs images generated with an unseen text prompt (``a dslr portrait of \textit{sks} person''). We further test the ASPL-disturbed DreamBooth models with different inference text prompts and get similar results. Fig. \ref{fig:prompts} provides some examples generated with these extra prompts.

\minisection{Textual Inversion \cite{gal2023an} and DreamBooth with LoRA \cite{hu2021lora}.} We additionally test ASPL againts other personalized text-to-image techniques. 
Textual Inversion learns new concepts by optimizing a word vector instead of finetuning the full model. LoRA uses low-rank weight updates to improve memory efficiency and is commonly used by the community. While being weaker than DreamBooth, they can be used to verify the robustness of our defense. As shown in Table \ref{tab:ti_lora}, ASPL successfully defends against both methods, further demonstrating our effectiveness against other personalization techniques.

\subsection{Adverse settings}
In this section, we investigate if our proposed defense still succeeds when some component is unknown, leading to a mismatch between the perturbation learning and the target DreamBooth model finetuning processes. Unless otherwise specified, all experiments will be conducted using ASPL with SD V2.1 and on the VGGFace2 dataset.

\minisection{Model mismatching.} The first scenario is when the pretrained generators are mismatched. We provide an example of transferring adversarial noise trained on SD v1.4 to defend DreamBooth models trained from v2.1 and v2.0 in the first and third rows in Table \ref{tab:cross}. ASPL still provides good scores as in Table \ref{tab:convenient}. We also examine the ensemble solution suggested in the literature, as discussed in Sec. \ref{sec:algorithms}. We combine that ensemble idea with ASPL, called E-ASPL, using SD v1.4, 1.5, and 2.1. It further improves the defense in both cases,
as illustrated in the upper half of Table \ref{tab:cross}.

\minisection{Term mismatching.} The malicious user can change the term representing the target from the default value (``sks'') to another, e.g., ``t@t''. As reported in the first row in the lower half of Table \ref{tab:cross}, this term mismatch has only a moderate effect on our results; key scores, like ISM, are still good.

\minisection{Prompt mismatching.} The malicious user can also use a different DreamBooth training prompt. ASPL still provides low ISM scores, as reported in the last row of Table \ref{tab:cross}.

\minisection{Image preprocessing.} We also evaluate the robustness of our defense under common image preprocessing techniques. As shown in Table \ref{tab:robustness_preprocessing}, applying Gaussian blur or JPEG compression on protected images slightly weakens the defense. However, the impact on generated image quality remains significant, as evidenced by the high BRISQUE scores across different settings. Hence, our defense demonstrates reasonable robustness against these techniques. 

\minisection{Real-world test.} Our method successfully disrupts personalized images generated by Astria \cite{astria}, a black-box commercial service (see in \cref{sec:real_test}).

\subsection{Uncontrolled settings}\label{sec:uncontrolled}
Anti-DreamBooth is designed for controlled settings, in which all images have protection noises added. In this section, we examine when the assumption does not hold, i.e., the malicious user gets some clean images of the target subject and mixes them with the perturbed images for DreamBooth training. Assuming the number of images for DreamBooth finetuning is fixed as 4, we examine three configurations with the number of clean images increasing from 1 to 3 (Table \ref{tab:uncontrolled}). Our defense is still effective when half of the images are perturbed, but its effectiveness reduces when more clean images are introduced. Still, these uncontrolled settings can be prevented if our system becomes popular and used by all social media with lawmakers' support.

\section{Conclusions}\label{sec:conclusions}
This paper reveals a potential threat of misused DreamBooth models and proposes a framework to counter it. Our solution is to perturb users' images with subtle adversarial noise so that any DreamBooth model trained on those images will produce poor personalized images. 
We designed several algorithms and extensively evaluated them. Our defense is effective, even in adverse conditions. 
In the future, we aim to improve the perturbation's imperceptibility and robustness \cite{cherepanova2021lowkey,yang2021defending,wang2022anti} and conquer uncontrolled settings.

{\small
\bibliographystyle{ieee_fullname}
\bibliography{egbib}

\begin{thebibliography}{10}\itemsep=-1pt

\bibitem{astria}
Astria.
\newblock \url{https://www.astria.ai/}.

\bibitem{celeba_id}
{CelebA-HQ-Face-Identity-and-Attributes-Recognition-PyTorch}.
\newblock
  \url{https://github.com/ndb796/CelebA-HQ-Face-Identity-and-Attributes-Recognition-PyTorch}.

\bibitem{Midjourney}
{Midjourney}.
\newblock \url{https://www.midjourney.com}.

\bibitem{SD}
{Stable Diffusion}.
\newblock \url{https://github.com/Stability-AI/stablediffusion}.

\bibitem{10.1145/3240323.3241729}
Fernando Amat, Ashok Chandrashekar, Tony Jebara, and Justin Basilico.
\newblock Artwork personalization at netflix.
\newblock In {\em Proceedings of the 12th ACM Conference on Recommender
  Systems}, RecSys '18, page 487–488, New York, NY, USA, 2018. Association
  for Computing Machinery.

\bibitem{square}
Maksym Andriushchenko, Francesco Croce, Nicolas Flammarion, and Matthias Hein.
\newblock Square attack: a query-efficient black-box adversarial attack via
  random search, 2019.

\bibitem{grad_obf}
Anish Athalye, Nicholas Carlini, and David Wagner.
\newblock Obfuscated gradients give a false sense of security: Circumventing
  defenses to adversarial examples, 2018.

\bibitem{eot}
Anish Athalye, Logan Engstrom, Andrew Ilyas, and Kevin Kwok.
\newblock Synthesizing robust adversarial examples, 2017.

\bibitem{balaji2022ediffi}
Yogesh Balaji, Seungjun Nah, Xun Huang, Arash Vahdat, Jiaming Song, Karsten
  Kreis, Miika Aittala, Timo Aila, Samuli Laine, Bryan Catanzaro, et~al.
\newblock ediff-i: Text-to-image diffusion models with an ensemble of expert
  denoisers.
\newblock {\em arXiv preprint arXiv:2211.01324}, 2022.

\bibitem{boundary}
Wieland Brendel, Jonas Rauber, and Matthias Bethge.
\newblock Decision-based adversarial attacks: Reliable attacks against
  black-box machine learning models, 2017.

\bibitem{Cao18}
Qiong Cao, Li Shen, Weidi Xie, Omkar~M. Parkhi, and Andrew Zisserman.
\newblock {VGGFace2}: A dataset for recognising faces across pose and age.
\newblock In {\em International Conference on Automatic Face and Gesture
  Recognition}, 2018.

\bibitem{cw}
N. Carlini and D. Wagner.
\newblock Towards evaluating the robustness of neural networks.
\newblock In {\em 2017 IEEE Symposium on Security and Privacy (SP)}, pages
  39--57, Los Alamitos, CA, USA, may 2017. IEEE Computer Society.

\bibitem{chang2023muse}
Huiwen Chang, Han Zhang, Jarred Barber, AJ Maschinot, Jose Lezama, Lu Jiang,
  Ming-Hsuan Yang, Kevin Murphy, William~T Freeman, Michael Rubinstein, et~al.
\newblock Muse: Text-to-image generation via masked generative transformers.
\newblock {\em arXiv preprint arXiv:2301.00704}, 2023.

\bibitem{hsja}
Jianbo Chen, Michael~I. Jordan, and Martin~J. Wainwright.
\newblock Hopskipjumpattack: A query-efficient decision-based attack, 2019.

\bibitem{cherepanova2021lowkey}
Valeriia Cherepanova, Micah Goldblum, Harrison Foley, Shiyuan Duan, John~P
  Dickerson, Gavin Taylor, and Tom Goldstein.
\newblock Lowkey: Leveraging adversarial attacks to protect social media users
  from facial recognition.
\newblock In {\em Proceedings of the International Conference on Learning
  Representations (ICLR)}, 2021.

\bibitem{autoattack}
Francesco Croce and Matthias Hein.
\newblock Reliable evaluation of adversarial robustness with an ensemble of
  diverse parameter-free attacks, 2020.

\bibitem{deng2020retinaface}
Jiankang Deng, Jia Guo, Evangelos Ververas, Irene Kotsia, and Stefanos
  Zafeiriou.
\newblock Retinaface: Single-shot multi-level face localisation in the wild.
\newblock In {\em Proceedings of the IEEE/CVF Conference on Computer Vision and
  Pattern Recognition}, pages 5203--5212, 2020.

\bibitem{deng2019arcface}
Jiankang Deng, Jia Guo, Niannan Xue, and Stefanos Zafeiriou.
\newblock Arcface: Additive angular margin loss for deep face recognition.
\newblock In {\em Proceedings of the IEEE/CVF conference on computer vision and
  pattern recognition}, pages 4690--4699, 2019.

\bibitem{dhariwal2021diffusion}
Prafulla Dhariwal and Alexander Nichol.
\newblock Diffusion models beat gans on image synthesis.
\newblock {\em Advances in Neural Information Processing Systems},
  34:8780--8794, 2021.

\bibitem{gal2023an}
Rinon Gal, Yuval Alaluf, Yuval Atzmon, Or Patashnik, Amit~Haim Bermano, Gal
  Chechik, and Daniel Cohen-or.
\newblock An image is worth one word: Personalizing text-to-image generation
  using textual inversion.
\newblock In {\em The Eleventh International Conference on Learning
  Representations}, 2023.

\bibitem{Goodfellow2014ExplainingAH}
Ian~J. Goodfellow, Jonathon Shlens, and Christian Szegedy.
\newblock Explaining and harnessing adversarial examples.
\newblock {\em CoRR}, abs/1412.6572, 2014.

\bibitem{SuPerFed}
Seok-Ju Hahn, Minwoo Jeong, and Junghye Lee.
\newblock Connecting low-loss subspace for personalized federated learning.
\newblock In {\em Proceedings of the 28th ACM SIGKDD Conference on Knowledge
  Discovery and Data Mining}, KDD '22, page 505–515, New York, NY, USA, 2022.
  Association for Computing Machinery.

\bibitem{ho2020denoising}
Jonathan Ho, Ajay Jain, and Pieter Abbeel.
\newblock Denoising diffusion probabilistic models.
\newblock In {\em Advances in neural information processing systems}, 2020.

\bibitem{hu2021lora}
Edward~J Hu, Phillip Wallis, Zeyuan Allen-Zhu, Yuanzhi Li, Shean Wang, Lu Wang,
  Weizhu Chen, et~al.
\newblock Lora: Low-rank adaptation of large language models.
\newblock In {\em International Conference on Learning Representations}, 2021.

\bibitem{hu2022protecting}
Shengshan Hu, Xiaogeng Liu, Yechao Zhang, Minghui Li, Leo~Yu Zhang, Hai Jin,
  and Libing Wu.
\newblock Protecting facial privacy: generating adversarial identity masks via
  style-robust makeup transfer.
\newblock In {\em Proceedings of the IEEE/CVF Conference on Computer Vision and
  Pattern Recognition}, pages 15014--15023, 2022.

\bibitem{huang2021initiative}
Qidong Huang, Jie Zhang, Wenbo Zhou, Weiming Zhang, and Nenghai Yu.
\newblock Initiative defense against facial manipulation.
\newblock In {\em Proceedings of the AAAI Conference on Artificial
  Intelligence}, volume~35, pages 1619--1627, 2021.

\bibitem{Ingram2022Dec}
David Ingram, Justine Goode, and Anjali Nair.
\newblock {You against the machine: Can you spot which image was created by
  A.I.?}
\newblock {\em NBC News}, Dec. 2022.

\bibitem{juefei2022countering}
Felix Juefei-Xu, Run Wang, Yihao Huang, Qing Guo, Lei Ma, and Yang Liu.
\newblock Countering malicious deepfakes: Survey, battleground, and horizon.
\newblock {\em International Journal of Computer Vision}, 130(7):1678--1734,
  2022.

\bibitem{karras2017progressive}
Tero Karras, Timo Aila, Samuli Laine, and Jaakko Lehtinen.
\newblock Progressive growing of gans for improved quality, stability, and
  variation.
\newblock {\em arXiv preprint arXiv:1710.10196}, 2017.

\bibitem{kumari2022customdiffusion}
Nupur Kumari, Bingliang Zhang, Richard Zhang, Eli Shechtman, and Jun-Yan Zhu.
\newblock Multi-concept customization of text-to-image diffusion.
\newblock {\em CVPR}, 2023.

\bibitem{bim}
Alexey Kurakin, Ian Goodfellow, and Samy Bengio.
\newblock Adversarial examples in the physical world, 2016.

\bibitem{li2023unganable}
Zheng Li, Ning Yu, Ahmed Salem, Michael Backes, Mario Fritz, and Yang Zhang.
\newblock Unganable: Defending against gan-based face manipulation.
\newblock In {\em USENIX Security}, 2023.

\bibitem{liang2023adversarial}
Chumeng Liang, Xiaoyu Wu, Yang Hua, Jiaru Zhang, Yiming Xue, Tao Song, Zhengui
  Xue, Ruhui Ma, and Haibing Guan.
\newblock Adversarial example does good: Preventing painting imitation from
  diffusion models via adversarial examples.
\newblock {\em arXiv preprint arXiv:2302.04578}, 2023.

\bibitem{liu2015faceattributes}
Ziwei Liu, Ping Luo, Xiaogang Wang, and Xiaoou Tang.
\newblock Deep learning face attributes in the wild.
\newblock In {\em Proceedings of International Conference on Computer Vision
  (ICCV)}, December 2015.

\bibitem{lllyasviel2023Apr}
lllyasviel.
\newblock {AdverseCleaner}, Apr. 2023.
\newblock [Online; accessed 3. Apr. 2023].

\bibitem{madry2018towards}
Aleksander Madry, Aleksandar Makelov, Ludwig Schmidt, Dimitris Tsipras, and
  Adrian Vladu.
\newblock Towards deep learning models resistant to adversarial attacks.
\newblock In {\em International Conference on Learning Representations}, 2018.

\bibitem{Brisque}
Anish Mittal, Anush~Krishna Moorthy, and Alan~Conrad Bovik.
\newblock No-reference image quality assessment in the spatial domain.
\newblock {\em IEEE Transactions on Image Processing}, 21(12):4695--4708, 2012.

\bibitem{deepfool}
S. Moosavi-Dezfooli, A. Fawzi, and P. Frossard.
\newblock Deepfool: A simple and accurate method to fool deep neural networks.
\newblock In {\em 2016 IEEE Conference on Computer Vision and Pattern
  Recognition (CVPR)}, pages 2574--2582, Los Alamitos, CA, USA, jun 2016. IEEE
  Computer Society.

\bibitem{Nichol2021GLIDETP}
Alex Nichol, Prafulla Dhariwal, Aditya Ramesh, Pranav Shyam, Pamela Mishkin,
  Bob McGrew, Ilya Sutskever, and Mark Chen.
\newblock Glide: Towards photorealistic image generation and editing with
  text-guided diffusion models.
\newblock In {\em International Conference on Machine Learning}, 2021.

\bibitem{radford2021learning}
Alec Radford, Jong~Wook Kim, Chris Hallacy, Aditya Ramesh, Gabriel Goh,
  Sandhini Agarwal, Girish Sastry, Amanda Askell, Pamela Mishkin, Jack Clark,
  et~al.
\newblock Learning transferable visual models from natural language
  supervision.
\newblock In {\em International conference on machine learning}, pages
  8748--8763. PMLR, 2021.

\bibitem{2020t5}
Colin Raffel, Noam Shazeer, Adam Roberts, Katherine Lee, Sharan Narang, Michael
  Matena, Yanqi Zhou, Wei Li, and Peter~J. Liu.
\newblock Exploring the limits of transfer learning with a unified text-to-text
  transformer.
\newblock {\em Journal of Machine Learning Research}, 21(140):1--67, 2020.

\bibitem{ramesh2022hierarchical}
Aditya Ramesh, Prafulla Dhariwal, Alex Nichol, Casey Chu, and Mark Chen.
\newblock Hierarchical text-conditional image generation with clip latents.
\newblock {\em arXiv preprint arXiv:2204.06125}, 2022.

\bibitem{Ramesh2022HierarchicalTI}
Aditya Ramesh, Prafulla Dhariwal, Alex Nichol, Casey Chu, and Mark Chen.
\newblock Hierarchical text-conditional image generation with clip latents.
\newblock {\em ArXiv}, abs/2204.06125, 2022.

\bibitem{rombach2022high}
Robin Rombach, Andreas Blattmann, Dominik Lorenz, Patrick Esser, and Bj{\"o}rn
  Ommer.
\newblock High-resolution image synthesis with latent diffusion models.
\newblock In {\em Proceedings of the IEEE/CVF Conference on Computer Vision and
  Pattern Recognition}, pages 10684--10695, 2022.

\bibitem{Roose2022Sep}
Kevin Roose.
\newblock {AI-Generated Art Won a Prize. Artists Aren{'}t Happy.}
\newblock {\em N.Y. Times}, Sept. 2022.

\bibitem{ruiz2020disrupting}
Nataniel Ruiz, Sarah~Adel Bargal, and Stan Sclaroff.
\newblock Disrupting deepfakes: Adversarial attacks against conditional image
  translation networks and facial manipulation systems.
\newblock In {\em Computer Vision--ECCV 2020 Workshops: Glasgow, UK, August
  23--28, 2020, Proceedings, Part IV 16}, pages 236--251. Springer, 2020.

\bibitem{ruiz2022dreambooth}
Nataniel Ruiz, Yuanzhen Li, Varun Jampani, Yael Pritch, Michael Rubinstein, and
  Kfir Aberman.
\newblock Dreambooth: Fine tuning text-to-image diffusion models for
  subject-driven generation.
\newblock 2022.

\bibitem{saharia2022photorealistic}
Chitwan Saharia, William Chan, Saurabh Saxena, Lala Li, Jay Whang, Emily
  Denton, Seyed Kamyar~Seyed Ghasemipour, Raphael Gontijo-Lopes, Burcu~Karagol
  Ayan, Tim Salimans, Jonathan Ho, David~J. Fleet, and Mohammad Norouzi.
\newblock Photorealistic text-to-image diffusion models with deep language
  understanding.
\newblock In Alice~H. Oh, Alekh Agarwal, Danielle Belgrave, and Kyunghyun Cho,
  editors, {\em Advances in Neural Information Processing Systems}, 2022.

\bibitem{sauer2023stylegan}
Axel Sauer, Tero Karras, Samuli Laine, Andreas Geiger, and Timo Aila.
\newblock Stylegan-t: Unlocking the power of gans for fast large-scale
  text-to-image synthesis.
\newblock {\em arXiv preprint arXiv:2301.09515}, 2023.

\bibitem{Schuhmann2022LAION5BAO}
Christoph Schuhmann, Romain Beaumont, Richard Vencu, Cade Gordon, Ross
  Wightman, Mehdi Cherti, Theo Coombes, Aarush Katta, Clayton Mullis, Mitchell
  Wortsman, Patrick Schramowski, Srivatsa Kundurthy, Katherine Crowson, Ludwig
  Schmidt, Robert Kaczmarczyk, and Jenia Jitsev.
\newblock Laion-5b: An open large-scale dataset for training next generation
  image-text models.
\newblock {\em ArXiv}, abs/2210.08402, 2022.

\bibitem{shamsian2021personalized}
Aviv Shamsian, Aviv Navon, Ethan Fetaya, and Gal Chechik.
\newblock Personalized federated learning using hypernetworks.
\newblock In {\em International Conference on Machine Learning}, pages
  9489--9502. PMLR, 2021.

\bibitem{shan2023glaze}
Shawn Shan, Jenna Cryan, Emily Wenger, Haitao Zheng, Rana Hanocka, and Ben~Y
  Zhao.
\newblock Glaze: Protecting artists from style mimicry by text-to-image models.
\newblock {\em arXiv preprint arXiv:2302.04222}, 2023.

\bibitem{shan2020fawkes}
Shawn Shan, Emily Wenger, Jiayun Zhang, Huiying Li, Haitao Zheng, and Ben~Y
  Zhao.
\newblock Fawkes: Protecting privacy against unauthorized deep learning models.
\newblock In {\em Proceedings of the 29th USENIX Security Symposium}, 2020.

\bibitem{sohl2015deep}
Jascha Sohl-Dickstein, Eric Weiss, Niru Maheswaranathan, and Surya Ganguli.
\newblock Deep unsupervised learning using nonequilibrium thermodynamics.
\newblock In {\em International Conference on Machine Learning}, 2015.

\bibitem{song2020denoising}
Jiaming Song, Chenlin Meng, and Stefano Ermon.
\newblock Denoising diffusion implicit models.
\newblock In {\em International Conference on Learning Representations}, 2021.

\bibitem{TerhorstKDKK20}
Philipp Terh{\"{o}}rst, Jan~Niklas Kolf, Naser Damer, Florian Kirchbuchner, and
  Arjan Kuijper.
\newblock {SER-FIQ:} unsupervised estimation of face image quality based on
  stochastic embedding robustness.
\newblock In {\em 2020 {IEEE/CVF} Conference on Computer Vision and Pattern
  Recognition, {CVPR} 2020, Seattle, WA, USA, June 13-19, 2020}, pages
  5650--5659. {IEEE}.

\bibitem{spsa}
Jonathan Uesato, Brendan O'Donoghue, Aaron van~den Oord, and Pushmeet Kohli.
\newblock Adversarial risk and the dangers of evaluating against weak attacks,
  2018.

\bibitem{von-platen-etal-2022-diffusers}
Patrick von Platen, Suraj Patil, Anton Lozhkov, Pedro Cuenca, Nathan Lambert,
  Kashif Rasul, Mishig Davaadorj, and Thomas Wolf.
\newblock Diffusers: State-of-the-art diffusion models.
\newblock \url{https://github.com/huggingface/diffusers}, 2022.

\bibitem{wang2022anti}
Run Wang, Ziheng Huang, Zhikai Chen, Li Liu, Jing Chen, and Lina Wang.
\newblock Anti-forgery: Towards a stealthy and robust deepfake disruption
  attack via adversarial perceptual-aware perturbations.
\newblock {\em arXiv preprint arXiv:2206.00477}, 2022.

\bibitem{yang2021defending}
Chaofei Yang, Leah Ding, Yiran Chen, and Hai Li.
\newblock Defending against gan-based deepfake attacks via transformation-aware
  adversarial faces.
\newblock In {\em 2021 international joint conference on neural networks
  (IJCNN)}, pages 1--8. IEEE, 2021.

\bibitem{yeh2020disrupting}
Chin-Yuan Yeh, Hsi-Wen Chen, Shang-Lun Tsai, and Sheng-De Wang.
\newblock Disrupting image-translation-based deepfake algorithms with
  adversarial attacks.
\newblock In {\em Proceedings of the IEEE/CVF Winter Conference on Applications
  of Computer Vision Workshops}, pages 53--62, 2020.

\bibitem{yu2022scaling}
Jiahui Yu, Yuanzhong Xu, Jing~Yu Koh, Thang Luong, Gunjan Baid, Zirui Wang,
  Vijay Vasudevan, Alexander Ku, Yinfei Yang, Burcu~Karagol Ayan, et~al.
\newblock Scaling autoregressive models for content-rich text-to-image
  generation.
\newblock {\em arXiv preprint arXiv:2206.10789}, 2022.

\bibitem{zhang2023adding}
Lvmin Zhang and Maneesh Agrawala.
\newblock Adding conditional control to text-to-image diffusion models.
\newblock {\em arXiv preprint arXiv:2302.05543}, 2023.

\bibitem{zhong2022opom}
Yaoyao Zhong and Weihong Deng.
\newblock Opom: Customized invisible cloak towards face privacy protection.
\newblock {\em IEEE Transactions on Pattern Analysis and Machine Intelligence},
  2022.

\end{thebibliography}
}

\appendix

\section{Additional quantitative results}
In the main paper, we comprehensively analyzed ASPL's performance on the VGGFace2 dataset. Here, we provide additional quantitative results on the CelebA-HQ dataset. We also report extra results with FSMG, the second-best defense algorithm, on the convenient settings.

\subsection{Ablation studies}

\minisection{Text-to-image generator version.} 
We investigate the effectiveness of our defense methods across different versions of SD models, including v1.4 and v1.5. 

As reported in \cref{tab:aspl_versions}, ASPL significantly decreases the identity scores (ISM) in CelebA-HQ, confirming its defense's effectiveness. Its scores, however, are not as good as in VGGFaces2. We can explain it by the fact that CelebA-HQ images are more constrained in pose and quality, reducing the diversity of the image set for DreamBooth and making their combined perturbation effect less severe.

As for FSMG, there is a similar pattern in all metrics on both VGGFace2 and CelebA-HQ, as presented in \cref{tab:fsmg_versions}. FSMG provides a slightly weaker defense compared with ASPL, confirming our observation in the main paper.

\vspace{2mm}
\minisection{Noise budget.} 
We further examine the impact of noise budget $\eta$ on FSMG and ASPL using SD v2.1 in \cref{tab:fsmg_noise,tab:aspl_noise}. As expected, increasing the noise budget leads to better defense scores, either with FSMG or ASPL and either in VGGFace2 or CelebA-HQ. Again, ASPL outperforms FSMG on most evaluation scores.

\minisection{Design choice for targeted defense.}
In the paper, we used a random subject from another dataset, guaranteeing a different identity. Here, we examine more advanced solutions. For each input image, we select the target from the outer dataset based on their identity distance (measured by ArcFace R50), which is either (1) the farthest or (2) in the 50-75 percentile (GLAZE's config). In both options, the targeted defenses show weak performance (Tab. \ref{tab:tbl_targeted}).

\subsection{Adverse settings}
In the main paper, we verified that our best protection method, i.e., ASPL, remained effective in VGGFace2 when some components of the target DreamBooth training were unknown, resulting in a disparity between the perturbation learning and the DreamBooth finetuning. Here we repeat those defense experiments but on the CelebA-HQ dataset to further confirm ASPL's effectiveness.

\minisection{Model mismatching.}
As can be seen in \cref{tab:aspl_cross}, the ASPL approach still works effectively on CelebA-HQ in the cross-model settings. Furthermore, the ensemble approach demonstrates a superior performance on all measurements, the same as the observation on VGGFace2.

\minisection{Term mismatching.} In realistic scenarios, the term representing the target in training DreamBooth might vary differently. To demonstrate this problem, we report ASPL's performance when the term ``sks'' is changed to ``t@t''. As can be seen in \cref{tab:aspl_cross}, our method still provides an extremely low ISM score, guaranteeing user protection regardless of the term mismatching.

\minisection{Prompt mismatching.}
This is the challenging setting when the attacker uses a prompt different from the one used in perturbation learning to train his/her DreamBooth model. In \cref{tab:aspl_cross}, though there is a drop in some metrics compared with the convenience settings, either the ISM or BRISQUE score remains relatively good.
This evidence further assures that our approaches are robust to the prompt mismatching problem.

\subsection{Uncontrolled settings}
We examine APSL in the uncontrolled settings on CelebaA-HQ (\cref{tab:aspl_uncontrolled}) and observe the same trend as reported on the VGGFace2 dataset.

\section{Real-world test.}
\label{sec:real_test}

In previous tests, we conducted experiments in laboratory mode. In this section, we examine if our proposed defense actually works in real-world scenarios by trying to disrupt personalized generation outputs of a black-box, commercialized AI service. We find Astria \cite{astria} satisfies our criteria and decide to use it in this test. Astria uses the basic DreamBooth setup that allows us to upload images of a specific target subject and input a generation prompt to acquire corresponding synthesized images. It also supports different model settings; we pick the recommended setting (SD v1.5 with face detection enabled) and a totally different one (Protogen 3.4 + Prism) for the tests.

\begin{figure*}
    \centering
    \includegraphics[width=\textwidth]{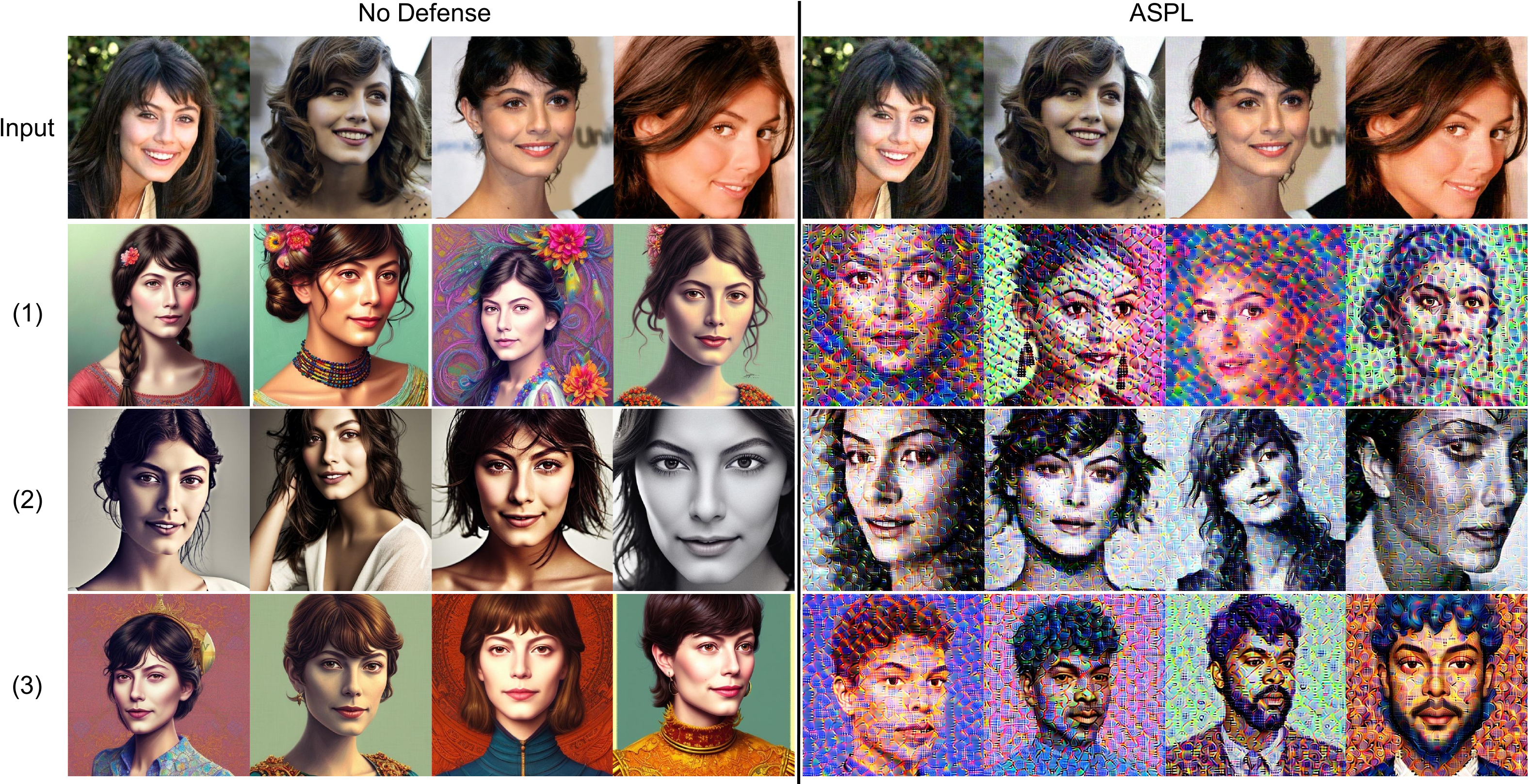}
    \caption{\textbf{Disrupting personalized images generated by Astria (SD v1.5 with face detection enabled)}. The prompts for image generation include: (1) ``portrait of \textit{sks} person portrait wearing fantastic Hand-dyed cotton clothes, embellished beaded feather decorative fringe knots, colorful pigtail, subtropical flowers and plants, symmetrical face, intricate, elegant, highly detailed, 8k, digital painting, trending on pinterest, harper's bazaar, concept art, sharp focus, illustration, by artgerm, Tom Bagshaw, Lawrence Alma-Tadema, greg rutkowski, alphonse Mucha'', (2) ``close up of face of \textit{sks} person fashion model in white feather clothes, official balmain editorial, dramatic lighting highly  detailed'', and (3) ``portrait of sks person prince :: by Martine Johanna and Simon Stålenhag and Chie Yoshii and Casey Weldon and wlop :: ornate, dynamic, particulate, rich colors, intricate, elegant, highly detailed, centered, artstation, smooth, sharp focus, octane render, 3d'' }
    \label{fig:astria1}
    \vspace{-5mm}
\end{figure*}

\begin{figure*}
    \centering
    \includegraphics[width=\textwidth]{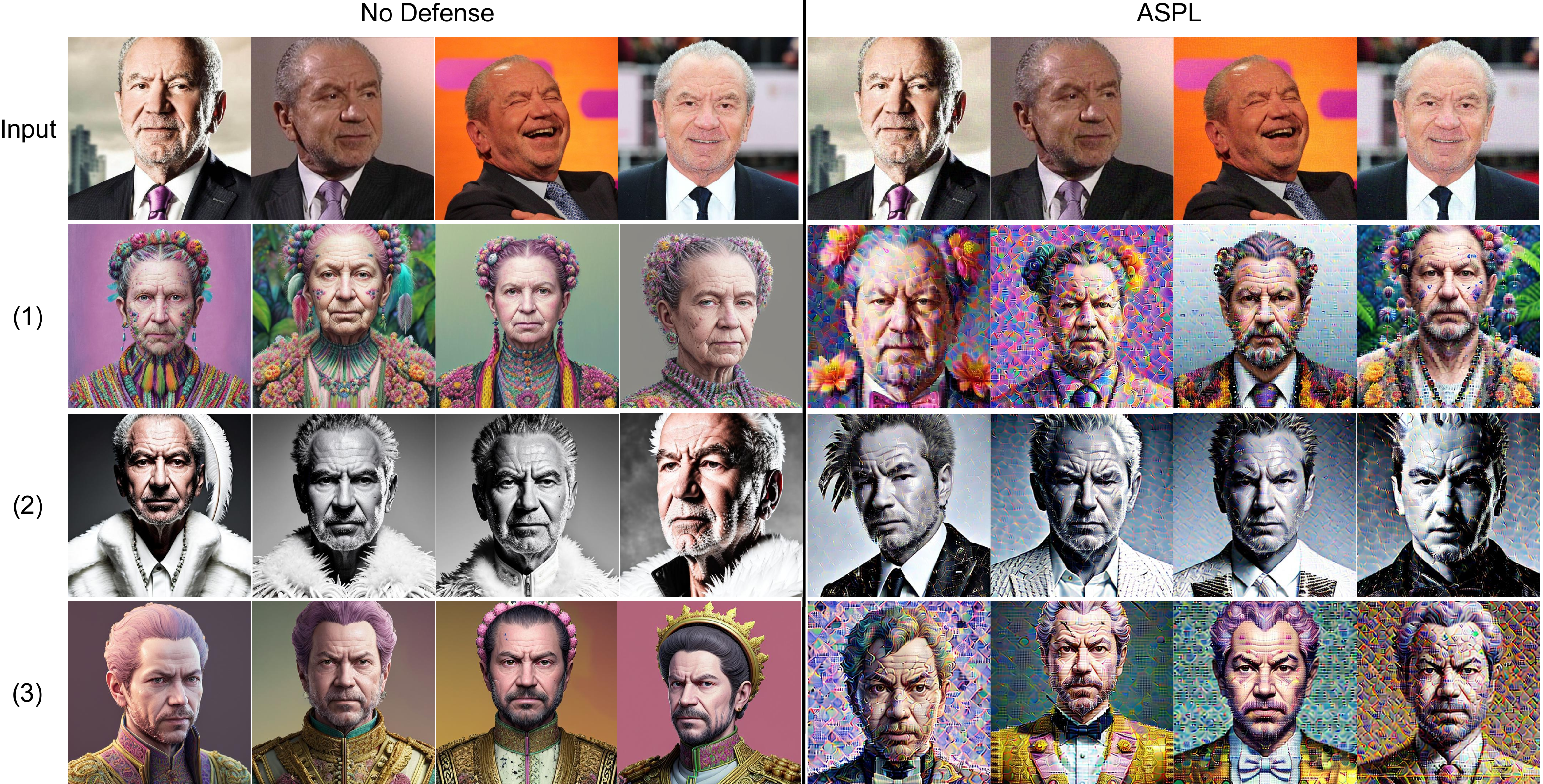}
    \caption{\textbf{Disrupting personalized images generated by Astria (Protogen with Prism and face detection enabled)}. The prompts for image generation include: (1) ``portrait of \textit{sks} person portrait wearing fantastic Hand-dyed cotton clothes, embellished beaded feather decorative fringe knots, colorful pigtail, subtropical flowers and plants, symmetrical face, intricate, elegant, highly detailed, 8k, digital painting, trending on pinterest, harper's bazaar, concept art, sharp focus, illustration, by artgerm, Tom Bagshaw, Lawrence Alma-Tadema, greg rutkowski, alphonse Mucha'', (2) ``close up of face of \textit{sks} person fashion model in white feather clothes, official balmain editorial, dramatic lighting highly  detailed'', and (3) ``portrait of sks person prince :: by Martine Johanna and Simon Stålenhag and Chie Yoshii and Casey Weldon and wlop :: ornate, dynamic, particulate, rich colors, intricate, elegant, highly detailed, centered, artstation, smooth, sharp focus, octane render, 3d'' }
    \label{fig:astria2}
    \vspace{-5mm}
\end{figure*}

We compare the output of Astria when using the original images and the adversarial images defended by our ASPL method with Stable Diffusion version 2.1 and $\eta=0.05$ in \cref{fig:astria1,fig:astria2}, using two different subjects and with each model setting, respectively. As can be seen, our method significantly reduces the quality of the generated images in various complex prompts and on both target models. Even though these services often rely on proprietary algorithms and architectures that are not transparent to the public, our method remains effective against them. This highlights the robustness of our approach, which can defend against these services without requiring knowledge of their underlying configurations.

\begin{table}[t]
\resizebox{\columnwidth}{!}{
\begin{tabular}{|l|l|l|l|l|}
\hline
 &
  \begin{tabular}[c]{@{}l@{}}FDFR$ \uparrow$\end{tabular} &
  \begin{tabular}[c]{@{}l@{}}ISM$\downarrow$\end{tabular} &
  \begin{tabular}[c]{@{}l@{}}SER-FQA$\downarrow$\end{tabular} &
  \begin{tabular}[c]{@{}l@{}}BRISQUE$\uparrow$\end{tabular} \\ \hline
ASPL                 &\textbf{0.76} &\textbf{0.28} &\textbf{0.30} &\textbf{39.00}  \\ \hline
Gaussian Blur K=3    &\textbf{0.59} &\textbf{0.34} &\textbf{0.39} &\textbf{42.67} \\
Gaussian Blur K=5    &0.36 &0.38 &0.57 &38.80 \\
Gaussian Blur K=7    &0.25 &0.45 &0.67 &32.76 \\
Gaussian Blur K=9    &0.24 &0.46 &0.71 &24.16 \\ \hline
JPEG Comp. Q=10      &0.19 &0.40 &0.69 &10.57 \\
JPEG Comp. Q=30      &0.20 &0.47 &0.71 &8.69  \\
JPEG Comp. Q=50      &0.18 &0.46 &0.73 &13.57 \\
JPEG Comp. Q=70      &\textbf{0.27} &\textbf{0.40} &\textbf{0.61} &\textbf{30.67} \\ \hline
No def., no preproc. &0.21 &0.48 &0.71 &9.64 \\ \hline
\end{tabular}
}
\caption{ASPL’s performance on VGGFace2, using the prompt ``\textit{a dslr portrait of \textit{sks} person}''.} 
\label{tab:robustness_preprocessing_b}
\end{table}

\section{Qualitative results}
We comprehensively analyzed our defense mechanism quantitatively in the main paper. Here, we provide additional qualitative results to back up those numbers and for visualization, as well.


\subsection{Ablation studies}

\begin{figure*}
    \centering
    \includegraphics[width=\textwidth]{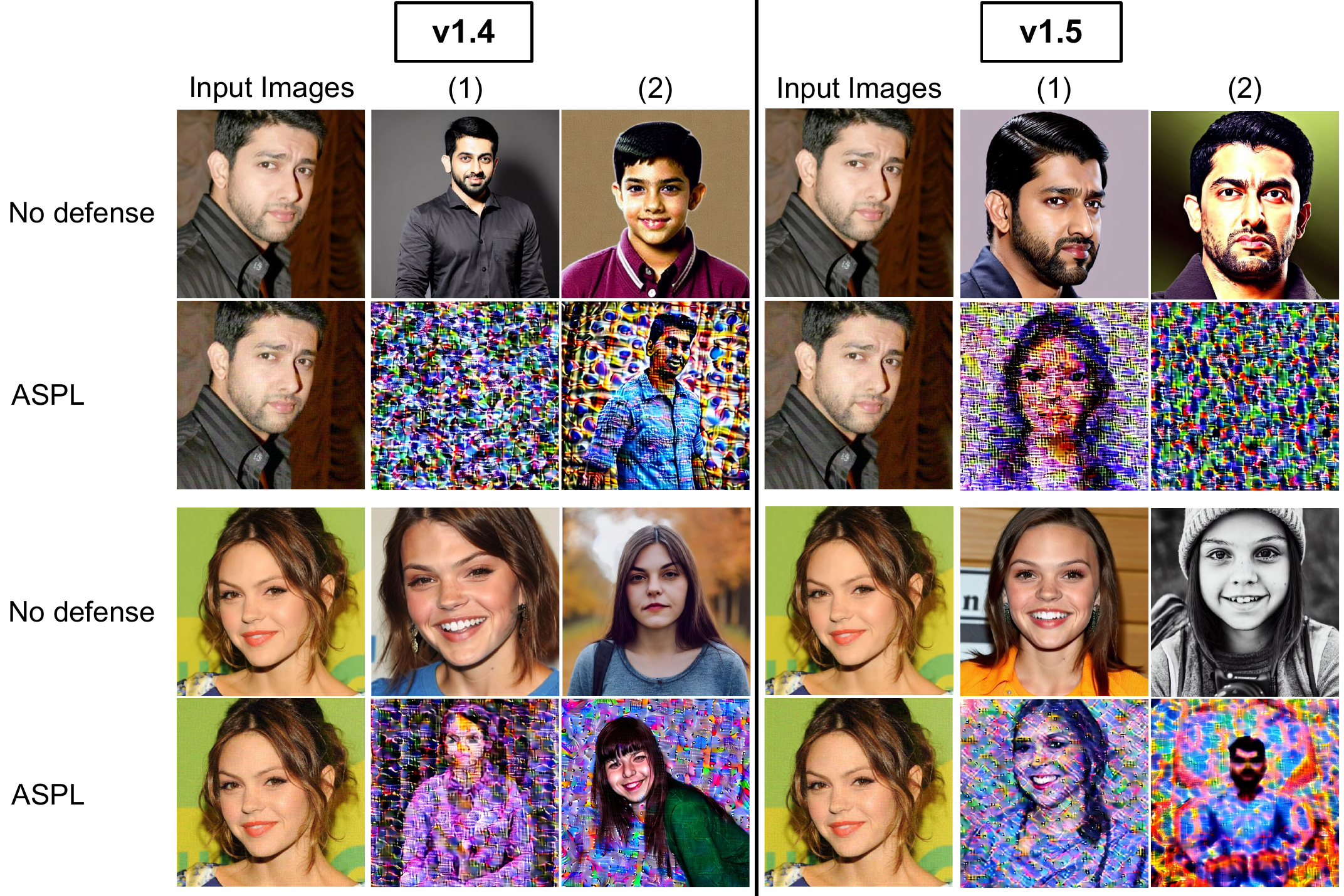}
    \caption{Qualitative results of ASPL with two different versions of SD models (v1.4 and v1.5) on VGGFace2. We provide in each test a single, representative input image. The generation prompts include (1) ``a photo of \textit{sks} person'' and (2) ``a dslr portrait of \textit{sks} person''.} 
    \label{fig:vgg_aspl_versions}
    \vspace{-5mm}
\end{figure*}

\begin{figure*}
    \centering
    \includegraphics[width=\textwidth]{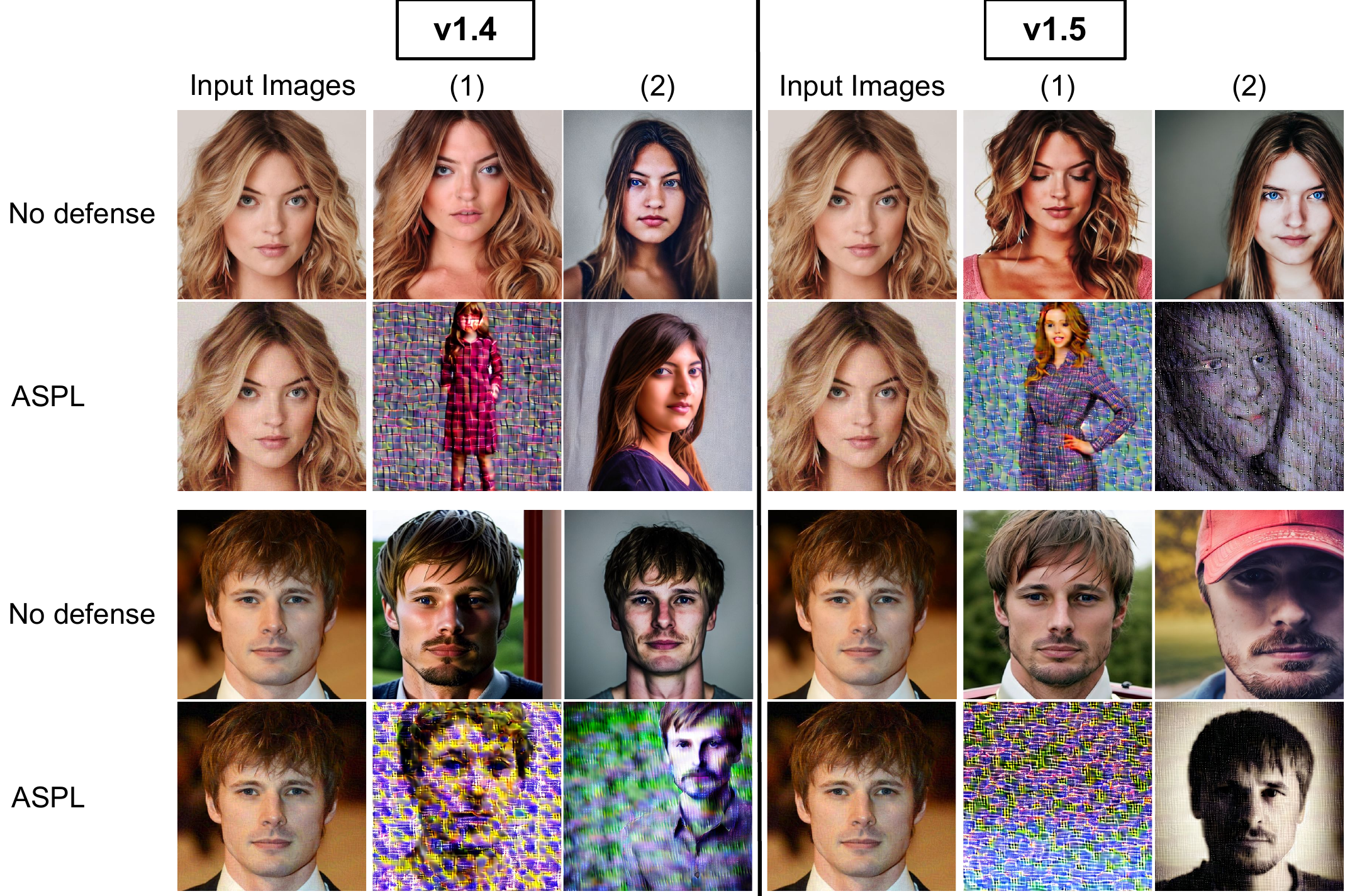}
    \caption{Qualitative results of ASPL with two different versions of SD models (v1.4 and v1.5) on CelebA-HQ. We provide in each test a single, representative input image. The generation prompts include (1) ``a photo of \textit{sks} person'' and (2) ``a dslr portrait of \textit{sks} person''.} 
    \label{fig:cel_aspl_versions}
    \vspace{-5mm}
\end{figure*}

\minisection{Text-to-image generator version.} 
We compare the defense performance of ASPL using two different versions of SD models (v1.4 and v1.5) on VGGFace2 in \cref{fig:vgg_aspl_versions}. The output images produced by both models with both prompts are strongly distorted with notable artifacts. We observe the same behavior in the corresponding experiments on CelebA-HQ, visualized in \cref{fig:cel_aspl_versions}.

\begin{figure*}
    \centering
    \includegraphics[width=\textwidth]{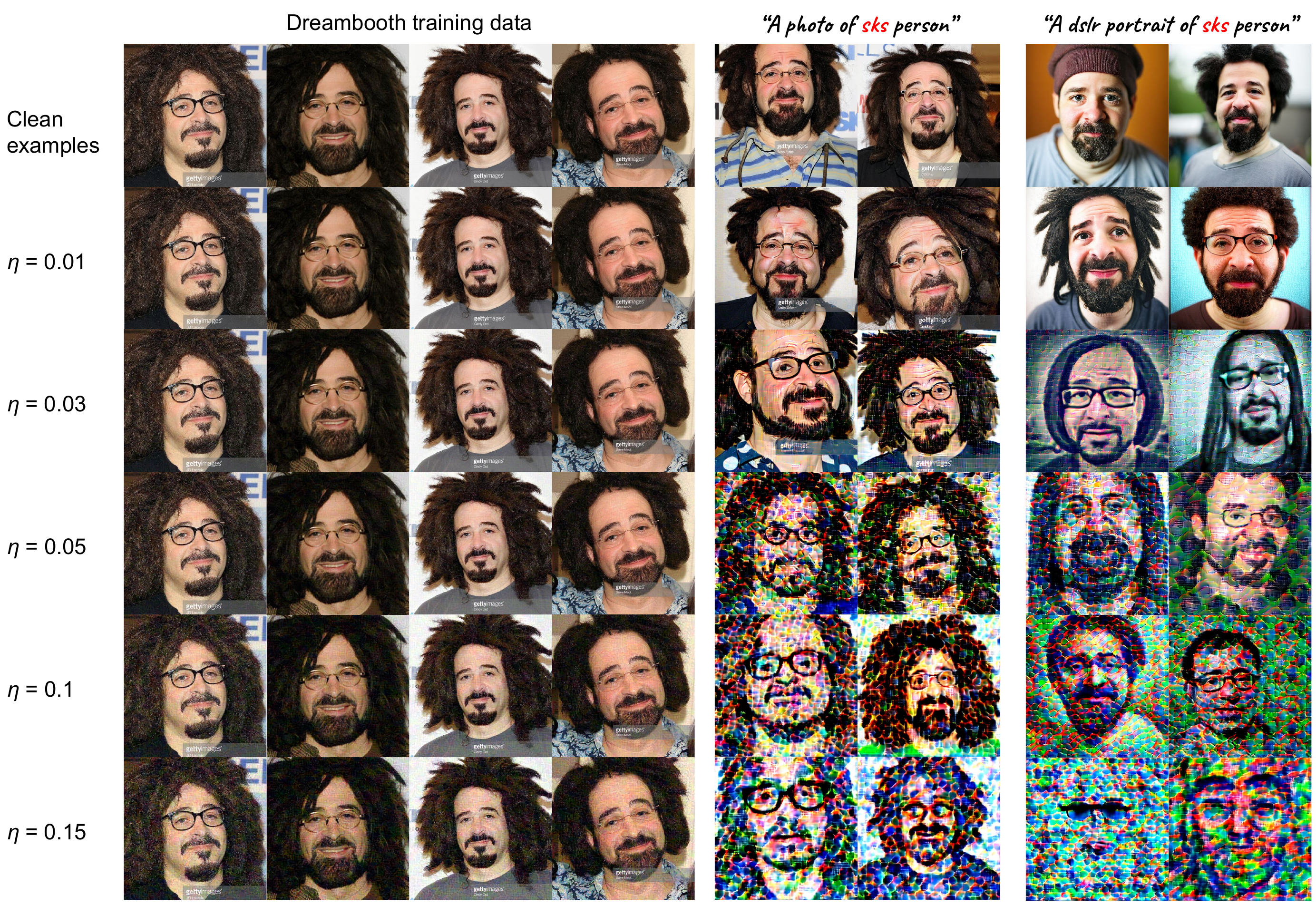}
    \caption{Qualativative results of ASPL with different noise budget on VGGFace2.} 
    \label{fig:noisebg_vgg}
    \vspace{-5mm}
\end{figure*}

\begin{figure*}
    \centering
    \includegraphics[width=\textwidth]{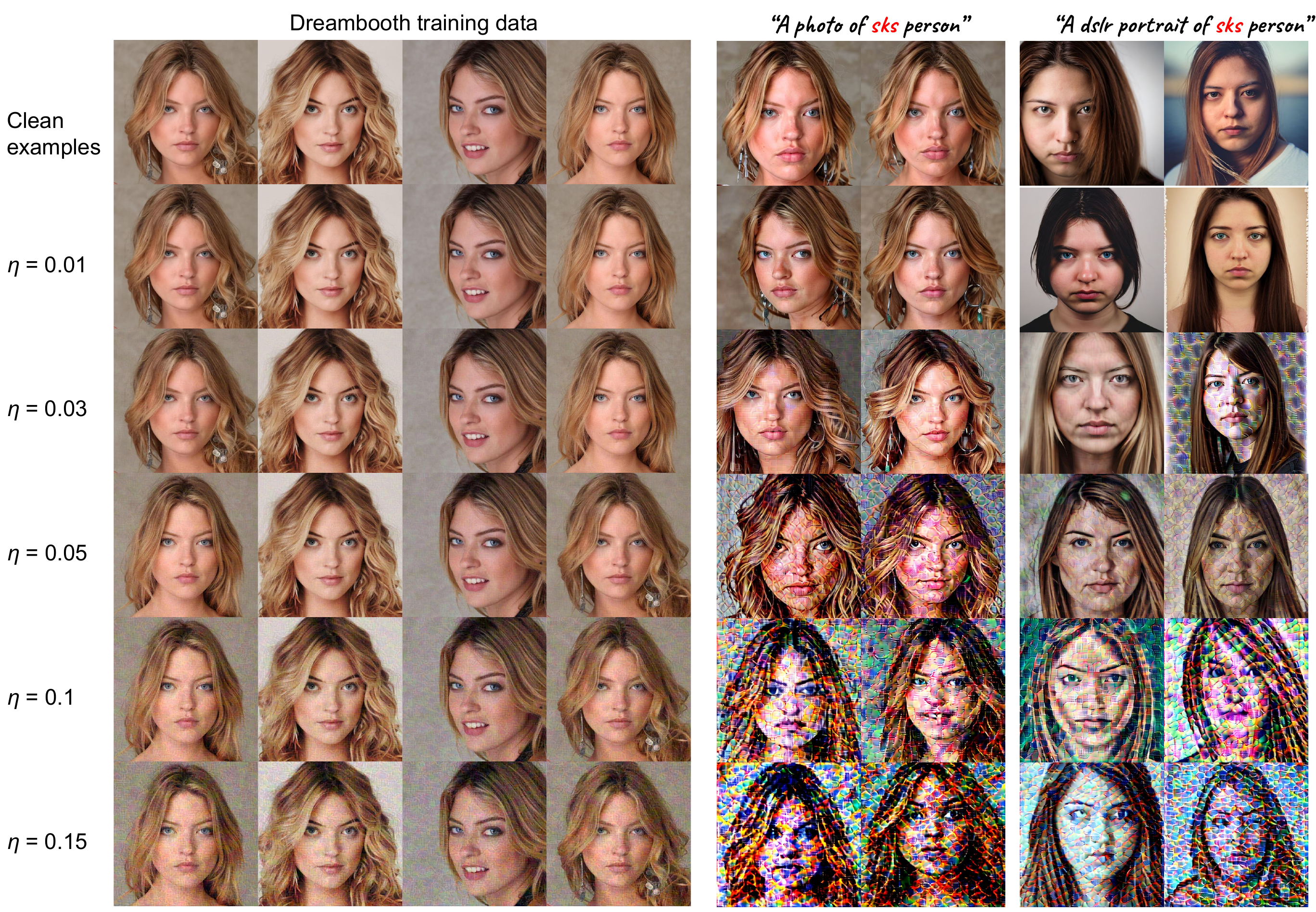}
    \caption{Qualitative results of ASPL with different noise budget on CelebA-HQ.} 
    \label{fig:noisebg_celeba}
    \vspace{-5mm}
\end{figure*}

\minisection{Noise budget.}
In order to better understand the impact of the noise budget, we present a grid of images for ASPL on VGGFace2 where the upper bound of noise's magnitude increases along the vertical axis in \cref{fig:noisebg_vgg}. It is evident that when the noise budget increases, the visibility of noise becomes more pronounced. Moreover, the allocated noise budget heavily influences the degree of output distortion, resulting in a trade-off between the visibility of noise in perturbed images and the level of output distortion. For further visulization on CelebA-HQ, please refer to \cref{fig:noisebg_celeba}.

\begin{table}[t]
    \centering
    \setlength{\tabcolsep}{3.5pt}
    \begin{tabular}{|l|c|c|c|c|}
\hline
Percentile & FDFR$\uparrow$ & ISM$\downarrow$ & SER-FQA$\downarrow$ & BRISQUE$\uparrow$  \\ \hline

Farthest      & 0.12           & 0.55            & 0.72                & \good{26.51} \\
50-75 &  0.12            & 0.55   &  0.71         &  22.19             \\
\hline
\end{tabular}%

    \vspace{-3mm}
    \caption{Defense performance of T-ASPL with  on CelebA-HQ in a convenient setting.}
    \label{tab:tbl_targeted}
    \vspace{-1mm}
\end{table}

\subsection{Adverse Setting}

\begin{figure*}
    \centering
    \includegraphics[width=\textwidth]{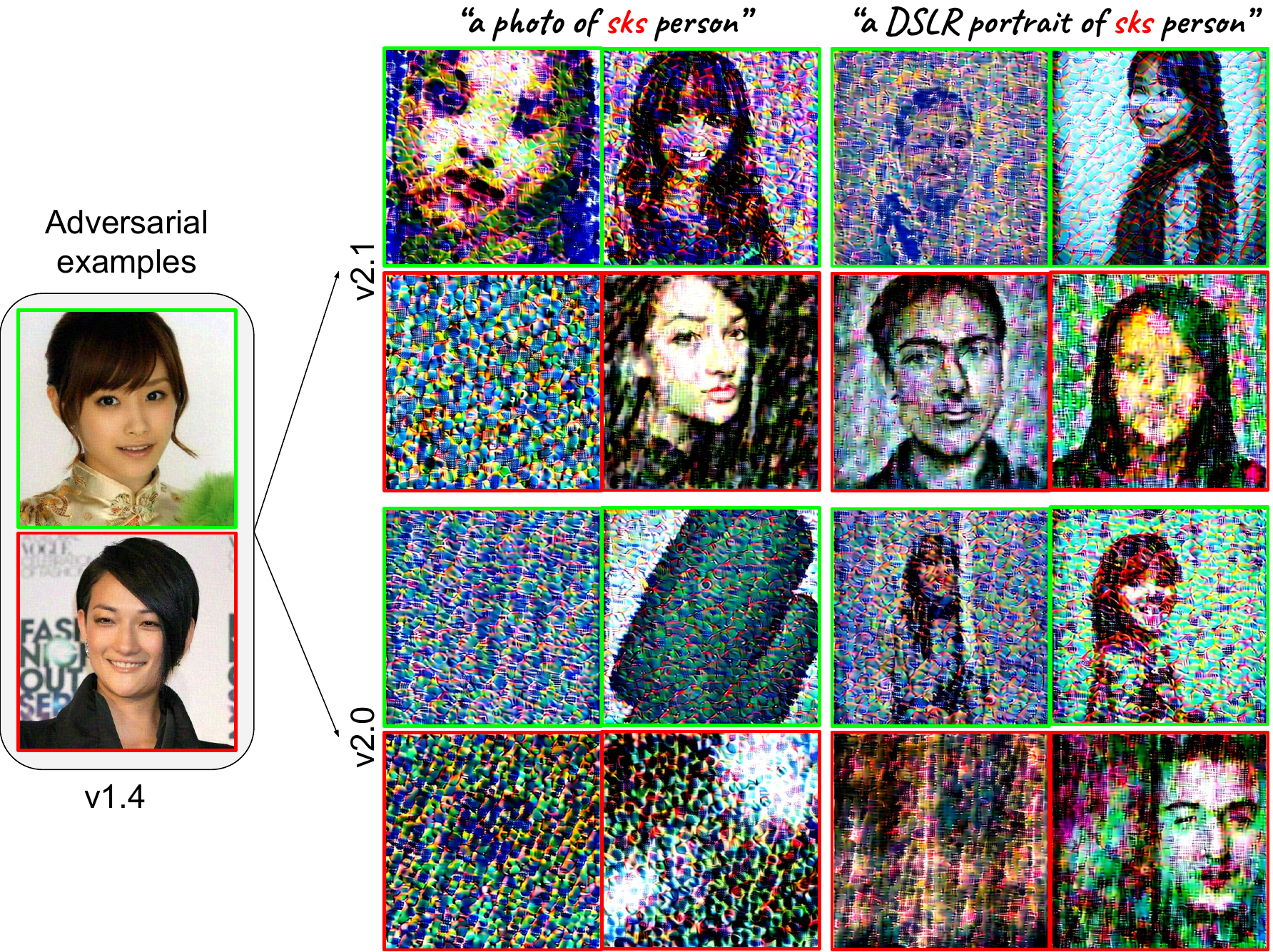}
    \caption{Qualitative results of ASPL in adverse settings on VGGFace2 where the SD model version in perturbation learning mismatches the one used in the DreamBooth finetuning stage (v1.4 $\rightarrow$ v2.1 and v1.4 $\rightarrow$ v2.0). We test with two random subjects and denote them in green and red, respectively.} 
    \label{fig:vgg_aspl_cross_model}
    \vspace{-5mm}
\end{figure*}

\minisection{Model mismatching.} 
In this section, we present the visual outputs of ASPL when a model mismatch occurs. Specifically, we train the image perturbation with SD v1.4, then use those images to disrupt DreamBooth models finetuned from v2.1 and v2.0, respectively. As illustrated in \cref{fig:vgg_aspl_cross_model}, our defense method is still effective in both cases, although transferring from v1.4 to v2.0 produces more noticeable artifacts than the previous scenario.

\begin{figure*}
    \centering
    \includegraphics[width=\textwidth]{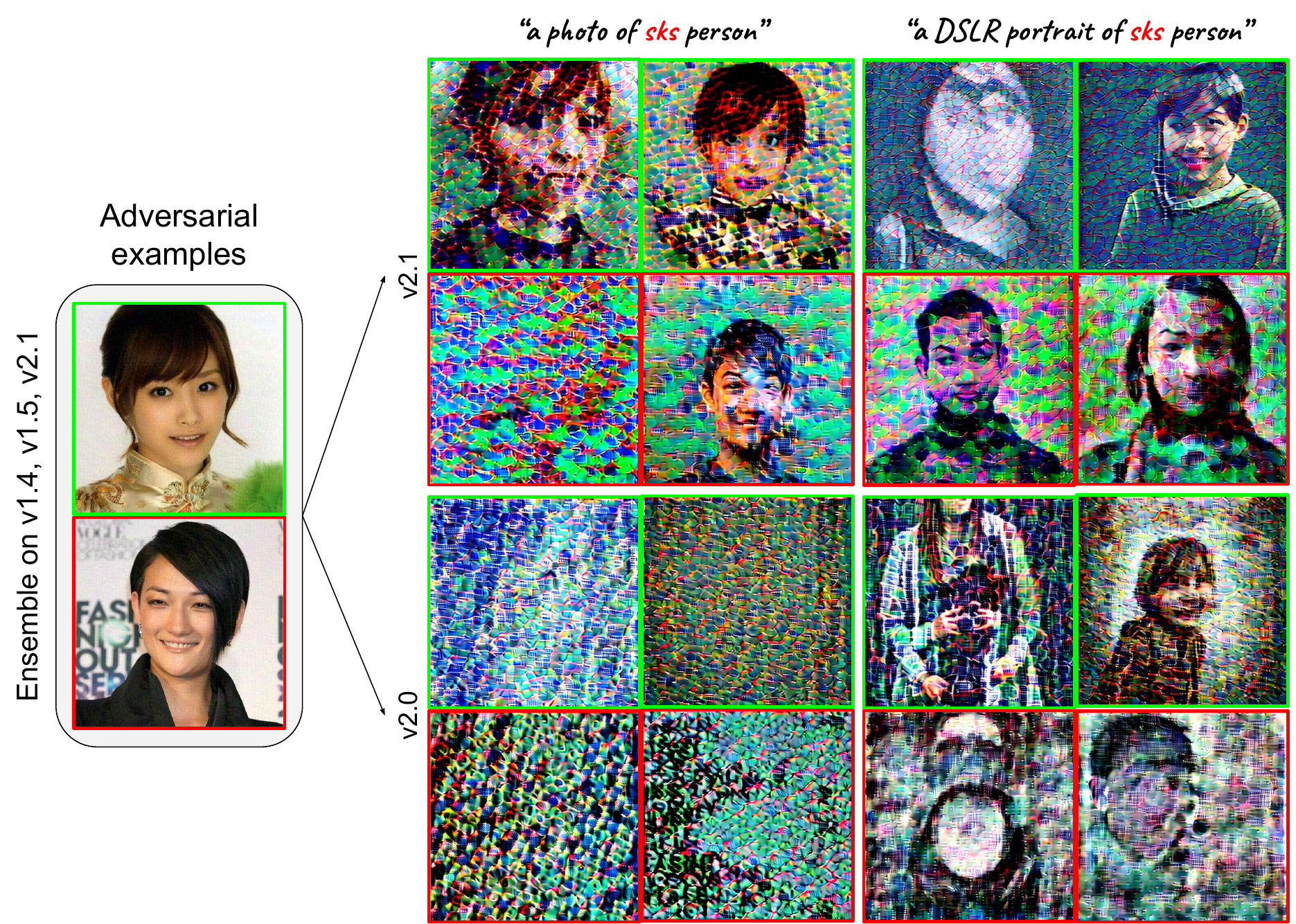}
    \caption{Qualitative results of E-ASPL on VGGFace2, where the ensemble model combines 3 versions of SD models, including v1.4, v1.5, and v2.1. Its performance is validated on two DreamBooth models finetuned on SD v2.1 and v2.0, respectively. We test with two random subjects and denote them in green and red, respectively.} 
    \label{fig:vgg_aspl_ensemble}
    \vspace{-5mm}
\end{figure*}

In addition to our primary analysis, our study provides qualitative results for E-ASPL, which employs an ensemble method to overcome the challenge of model mismatching. Specifically, we combined knowledge from three versions of SD models (v1.4, v1.5, and v2.1). The results, illustrated in \cref{fig:vgg_aspl_ensemble}, demonstrate the superior performance of E-ASPL in countering model mismatching where most images are heavily distorted.

\begin{figure*}
    \centering
    \includegraphics[width=\textwidth]{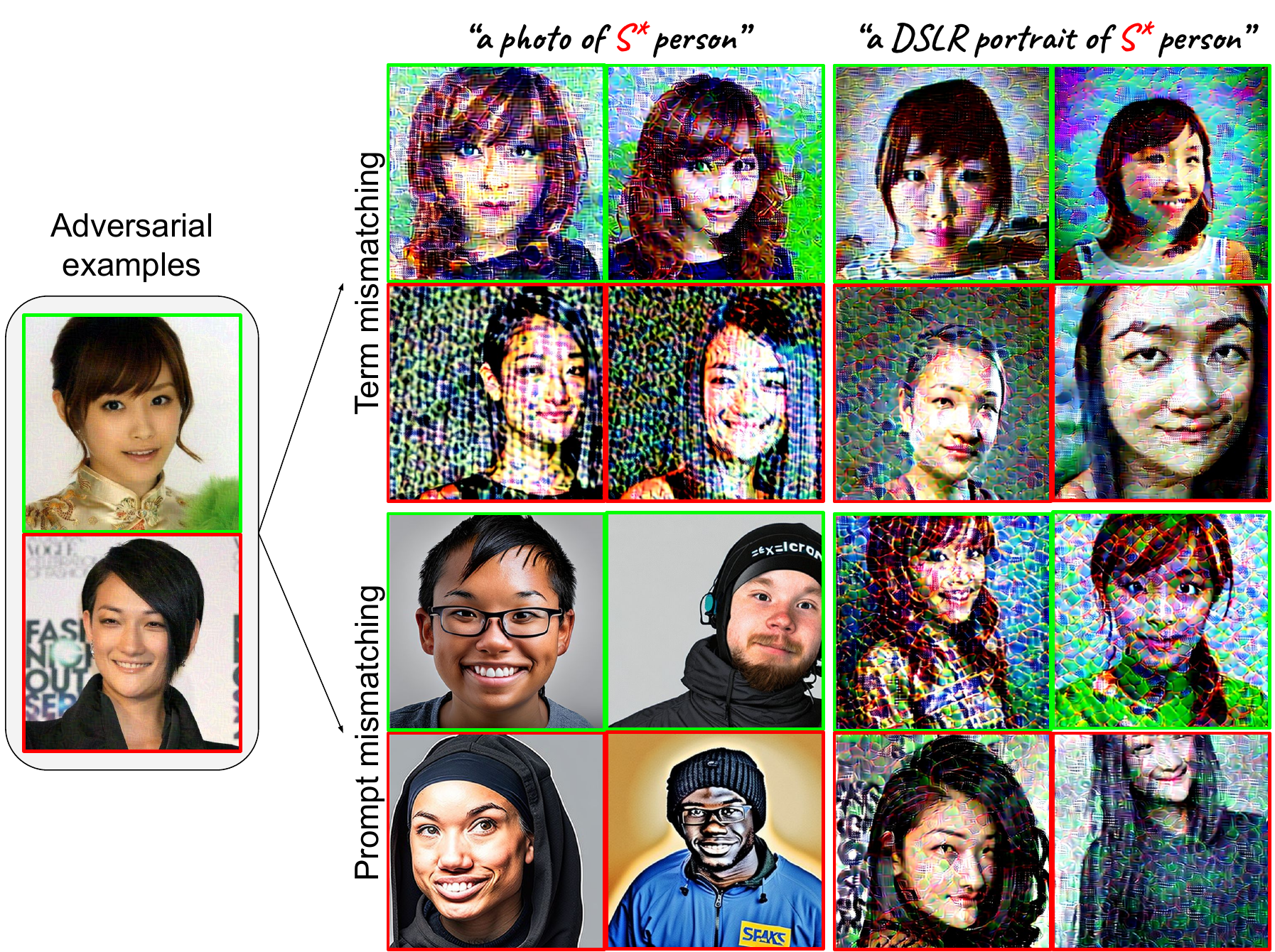}
    \caption{Qualitative results of ASPL on VGGFace2 where the training term and prompt of the target DreamBooth model mismatch the ones in perturbation learning. In the first scenario, the training term is changed from ``sks'' to ``t@t''. In the second scenario, the training prompt is replaced with ``a DSLR portrait of \textit{sks} person'' instead of ``a photo of \textit{sks} person''. Here, \textit{$S^*$} is ``t@t'' for term mismatching and ``sks'' for prompt mismatching. We test with two random subjects and denote them in green and red, respectively.} 
    \label{fig:vgg_aspl_cterm_cprompt}
    \vspace{-5mm}
\end{figure*}

\minisection{Term mismatching.}
Despite the discrepancy of term replacement (from ``sks'' to ``t@t''), ASPL still demonstrates its effectiveness on two provided subjects and two provided prompts (as in \cref{fig:vgg_aspl_cterm_cprompt}). However, the change in the training term may result in slightly weaker artifacts compared to the original setting.

\minisection{Prompt mismatching.} 
The results depicted in \cref{fig:vgg_aspl_cterm_cprompt} indicate that the finetuning of the DreamBooth model with various prompts, such as "a DSLR portrait of \textit{sks} person", can impact the degree of output distortion to some extent. It is important to note that prompt mismatching can alter the behavior of our defense method on a different prompt, such as "a photo of \textit{sks} person", which can change the identity of the target subject in the generated images.

\subsection{Uncontrolled settings.} 
\begin{figure*}
    \centering
    \includegraphics[width=\textwidth]{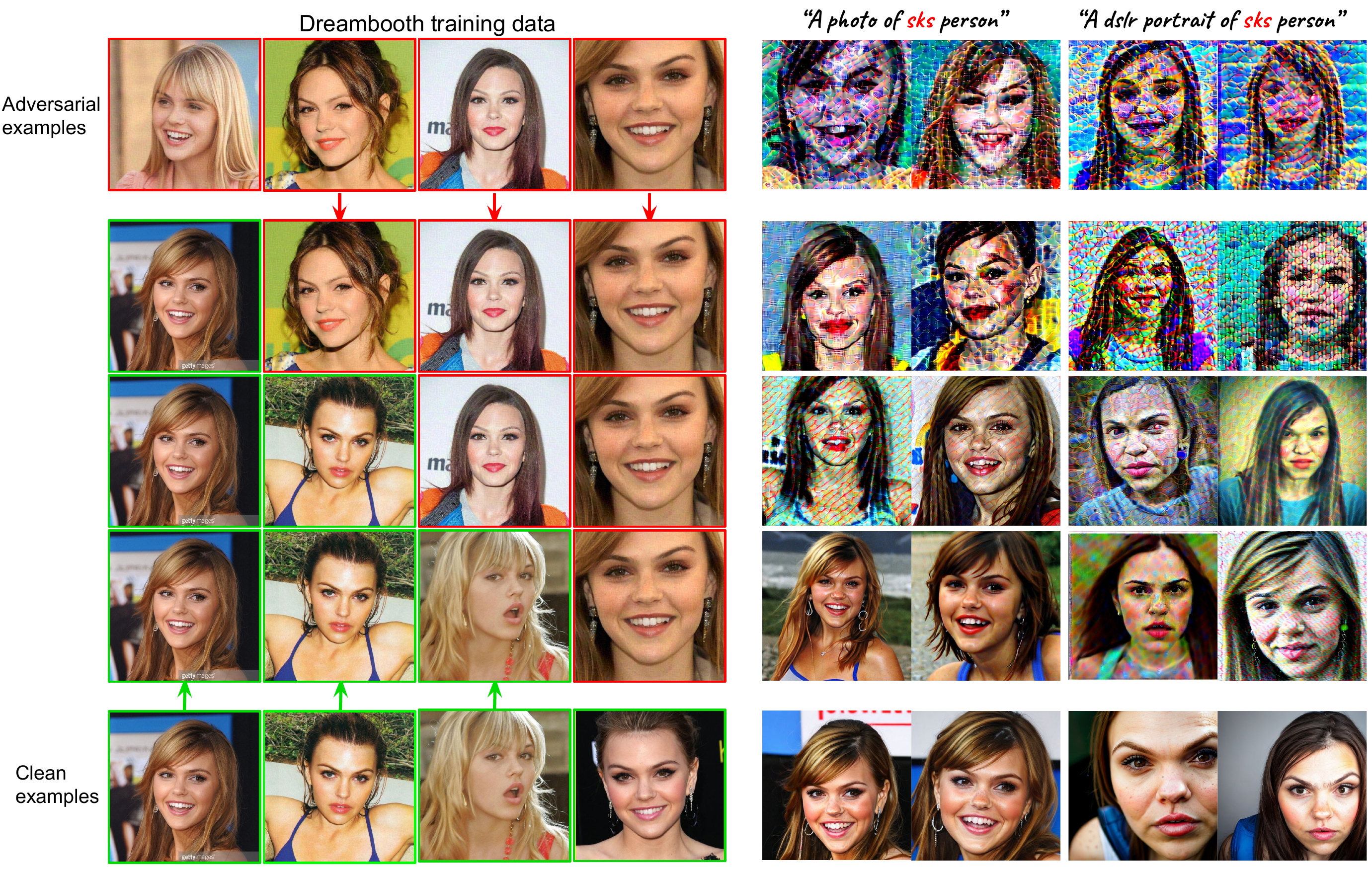}
    \caption{Qualitivative results of ASPL in uncontrolled setting on VGGFace2. We denote the perturbed examples and the leaked clean examples in red and green, respectively.} 
    \label{fig:uncontrolled_vgg}
    \vspace{-5mm}
\end{figure*}

All previous results are for controlled settings, in which we have access to all images needing protection. Here, we also include some qualitative results for uncontrolled settings where a mixture of clean and perturbed images are used for finetuning DreamBooth. We use the same settings as the one in the main paper, with the number of images for DreamBooth being fixed at 4 and the number of clean images gradually increase. As can be seen in \cref{fig:uncontrolled_vgg}, our method is more effective when more perturbed data are used and vice versa.

\section{Robustness Evaluation}
\label{sec:robust_extra}
For the purpose of straightforward comparisons, we use our default configuration which is ASPL with $\eta=0.05$ on all of our experiments in the main paper. Here, we further present quantitative results for a different prompt (\cref{tab:robustness_preprocessing_b}) rather than the one used for training in the main paper, along with corresponding qualitative figures for better demonstration.

\subsection{Robustness to Image Compression}
\begin{figure*}
    \centering
    \includegraphics[width=\textwidth]{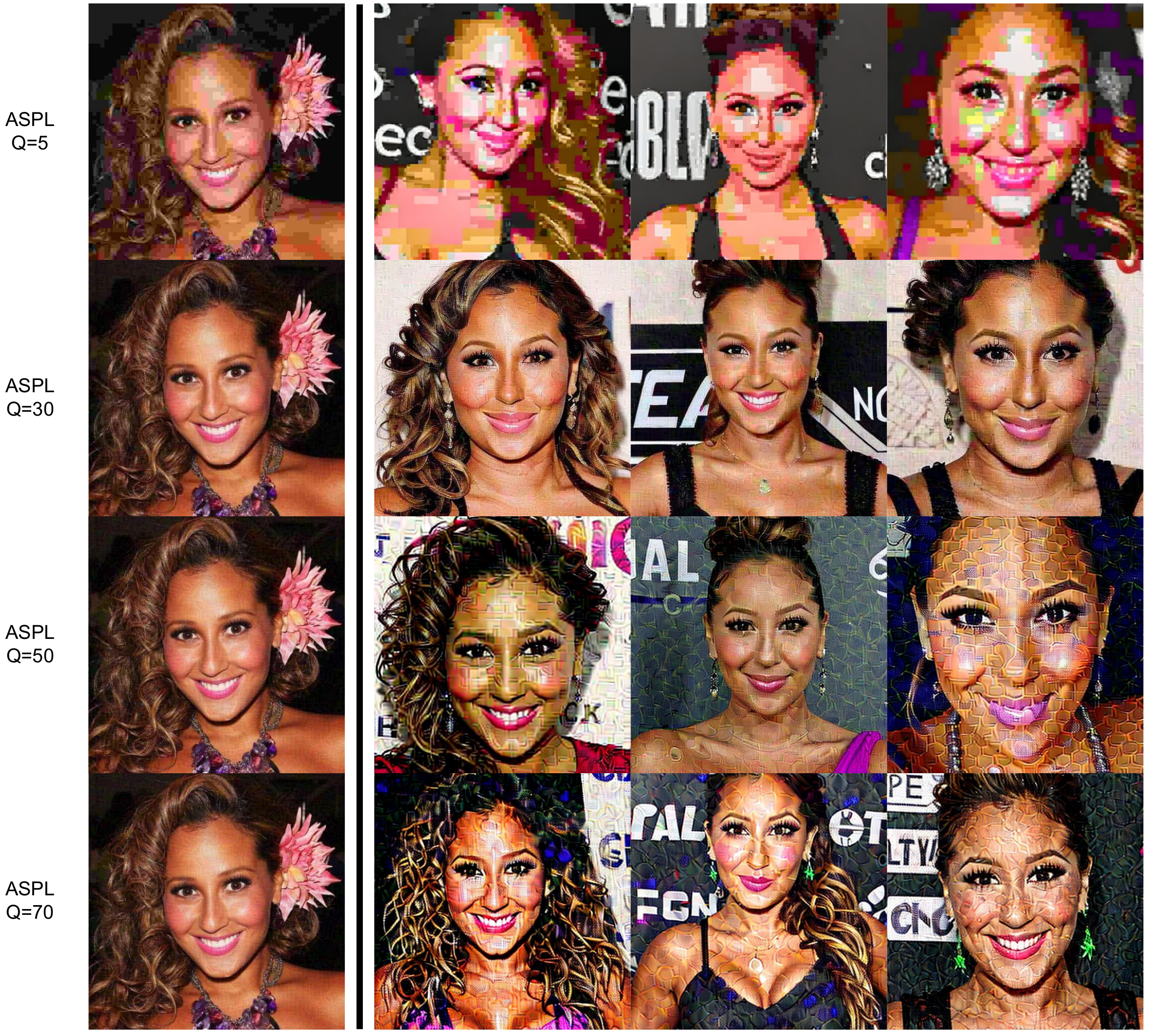}
    \caption{Qualitative results of ASPL under different JPEG Compression level.}
    \label{fig:robust_jpeg}
    \vspace{-5mm}
\end{figure*}

In \cref{fig:robust_jpeg}, the results revealed that our defense remained effective with significant artifacts in the generated images at compression levels of $50\%$ and above. Even though most adversarial artifacts were removed under $30\%$ compression, low-quality training images resulting from a low compression rate still caused significant image degradation.

\subsection{Gaussian Smoothing}
\begin{figure*}
    \centering
    \includegraphics[width=\textwidth]{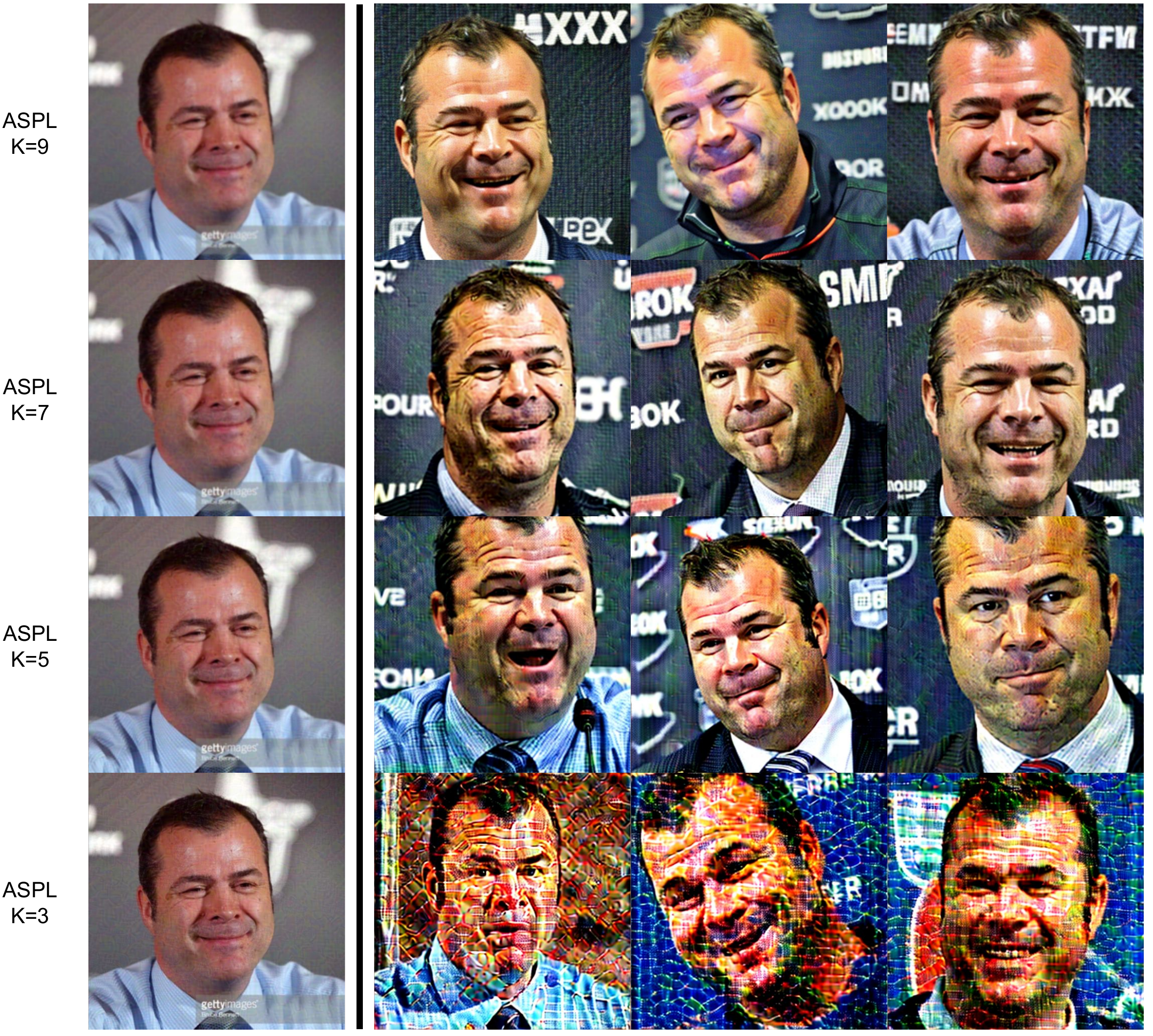}
    \caption{Qualitative results of ASPL after applying Gaussian Blur.}
    \label{fig:robust_gaussian}
    \vspace{-5mm}
\end{figure*}

 As can be seen in \cref{fig:robust_gaussian}, our defense remains effective with perceivable artifacts until kernel size 7. Although the artifacts produced by our defense method were lessened when increasing the kernel size, the quality of the generated images also degraded, as the generated images become blurry with unnatural background.

\subsection{Robustness to Adverse Cleaner}

\begin{figure*}
    \centering
    \includegraphics[width=\textwidth]{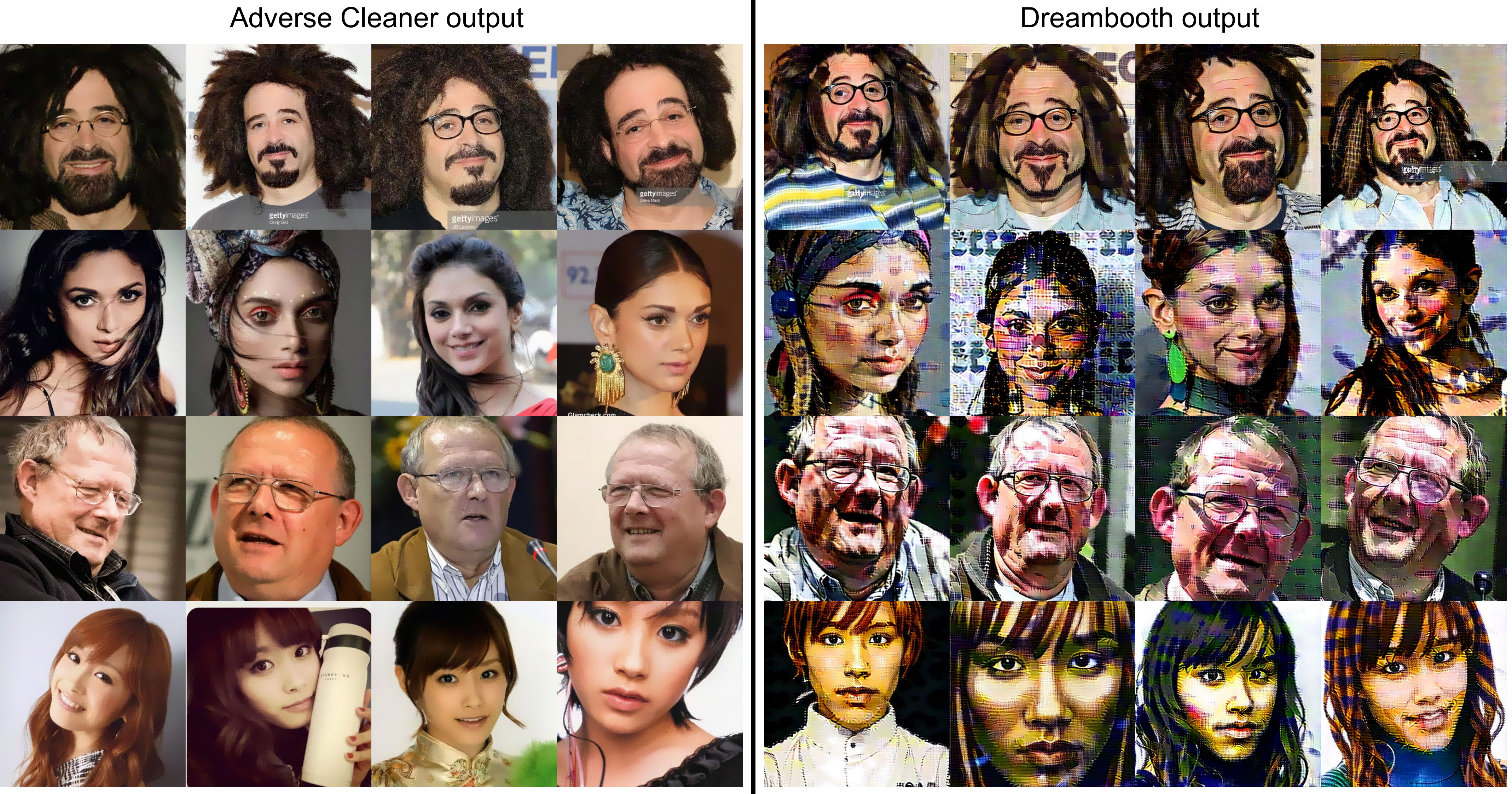}
    \caption{Qualitative results of ASPL under Adverse Cleaner.}
    \label{fig:robust_adverse_cleaner}
    \vspace{-5mm}
\end{figure*}

Finally, we tested our method against the Adverse Cleaner \cite{lllyasviel2023Apr}, which is currently the go-to library for cleaning images. We report the qualitative result in \cref{fig:robust_adverse_cleaner}.  Notably, our defense method demonstrated its robustness even against the smoothing effect of denoised images.

\subsection{Discussion}
Initial test results indicate that our defense scheme is quite robust against simple image processing methods such as Gaussian Blur and JPEG compression. However, we acknowledge that these processing methods may not represent all possible techniques to erase the protection. Thus, we plan to explore additional algorithms and techniques to further improve our defense system's robustness in future works.

\begin{table*}[t]
    \centering
    \setlength{\tabcolsep}{3.5pt}
    \begin{tabular}{l|c|cccc|cccc}
        \specialrule{.09em}{.04em}{.04em} 
         \multirow{2}{*}{Version} & \multirow{2}{*}{Defense?} & \multicolumn{4}{|c|}{``a photo of \textit{sks} person''} & \multicolumn{4}{c}{``a dslr portrait of \textit{sks} person''}\\
         \cline{3-10}
          &  & FDFR$\uparrow$ & ISM$\downarrow$ & SER-FQA$\downarrow$ & BRISQUE$\uparrow$ & FDFR$\uparrow$ & ISM$\downarrow$ & SER-FQA$\downarrow$ & BRISQUE$\uparrow$ \\
         \hline
         \multirow{2}{*}{v1.4} & \xmark & 0.07 & 0.48 &0.66 &16.09 & \textbf{0.11} & 0.40 &0.67 &10.31 \\
          & \cmark & \textbf{0.28} & \textbf{0.29} & \textbf{0.47} & \textbf{20.05} & 0.06 & \textbf{0.31} & \textbf{0.64} & \textbf{10.55} \\
          \hline
         \multirow{2}{*}{v1.5} & \xmark & 0.06 & 0.53 &0.69 &14.45 & \textbf{0.07} & 0.39 &0.68 &8.95 \\
          & \cmark & \textbf{0.16} & \textbf{0.36} & \textbf{0.58} & \textbf{21.09} &0.06 & \textbf{0.26} & \textbf{0.64} & \textbf{12.28} \\
        \specialrule{.09em}{.04em}{.04em} 
    \end{tabular}
    \vspace{-3mm}
    \caption{Defense performance of ASPL with different generator versions on CelebA-HQ in a convenient setting.}
    \label{tab:aspl_versions}
    \vspace{-1mm}
\end{table*}

\begin{table*}[t]
    \centering
    \setlength{\tabcolsep}{3.5pt}
    \begin{tabular}{l|c|cccc|cccc}
        \toprule
        \multicolumn{10}{c}{\textbf{VGGFace2}} \\
        \specialrule{.09em}{.04em}{.04em} 
         \multirow{2}{*}{Version} & \multirow{2}{*}{Defense?} & \multicolumn{4}{|c|}{``a photo of \textit{sks} person''} & \multicolumn{4}{c}{``a dslr portrait of \textit{sks} person''}\\
         \cline{3-10}
          &  & FDFR$\uparrow$ & ISM$\downarrow$ & SER-FQA$\downarrow$ & BRISQUE$\uparrow$ & FDFR$\uparrow$ & ISM$\downarrow$ & SER-FQA$\downarrow$ & BRISQUE$\uparrow$ \\
         \hline
         \multirow{2}{*}{v1.4} & \xmark & 0.05 & 0.46 &0.65 &21.06 & 0.08 & 0.44 &0.64 &10.05 \\
          & \cmark & \textbf{0.73} & \textbf{0.21} & \textbf{0.17} & \textbf{25.88} & \textbf{0.13} & \textbf{0.28} & \textbf{0.57} & \textbf{13.46} \\
          \hline
         \multirow{2}{*}{v1.5} & \xmark & 0.07 & 0.49 &0.65 &18.53 & 0.07 & 0.45 &0.64 &10.57 \\
          & \cmark & \textbf{0.61} & \textbf{0.21} & \textbf{0.26} & \textbf{23.89} & \textbf{0.11} & \textbf{0.26} & \textbf{0.57} & \textbf{18.00} \\

        \toprule
        \multicolumn{10}{c}{\textbf{CelebA-HQ}} \\
        \specialrule{.09em}{.04em}{.04em} 
         \multirow{2}{*}{Version} & \multirow{2}{*}{Defense?} & \multicolumn{4}{|c|}{``a photo of \textit{sks} person''} & \multicolumn{4}{c}{``a dslr portrait of \textit{sks} person''}\\
         \cline{3-10}
          &  & FDFR$\uparrow$ & ISM$\downarrow$ & SER-FQA$\downarrow$ & BRISQUE$\uparrow$ & FDFR$\uparrow$ & ISM$\downarrow$ & SER-FQA$\downarrow$ & BRISQUE$\uparrow$ \\
         \hline
         \multirow{2}{*}{v1.4} & \xmark & 0.07 & 0.48 &0.66 &16.09 & \textbf{0.11} & 0.40 &0.67 &10.31 \\
          & \cmark & \textbf{0.29} & \textbf{0.32} & \textbf{0.48} & \textbf{20.83} & 0.07 & \textbf{0.29} & \textbf{0.63} & \textbf{12.00} \\
          \hline
         \multirow{2}{*}{v1.5} & \xmark & 0.06 & 0.53 &0.69 &14.45 & \textbf{0.07} & 0.39 & 0.68 &8.95 \\
          & \cmark & \textbf{0.13} & \textbf{0.38} & \textbf{0.60} & \textbf{20.43} & 0.06 & \textbf{0.28} & \textbf{0.65} & \textbf{13.27} \\
        \bottomrule
        
    \end{tabular}
    \vspace{-3mm}
    \caption{Defense performance of FSMG with different generator versions on VGGFace2 and CelebA-HQ in a convenient setting.}
    \label{tab:fsmg_versions}
    \vspace{-1mm}
\end{table*}

\begin{table*}[t]
    \centering
    \setlength{\tabcolsep}{3pt}
    \begin{tabular}{c|cccc|cccc}
        \toprule
        \multicolumn{9}{c}{\textbf{VGGFace2}} \\
        \midrule
         \multirow{2}{*}{$\eta$} & \multicolumn{4}{|c|}{``a photo of \textit{sks} person''} & \multicolumn{4}{c}{``a dslr portrait of \textit{sks} person''}\\
         \cline{2-9}
          & FDFR$\uparrow$ & ISM$\downarrow$ & SER-FQA$\downarrow$ & BRISQUE$\uparrow$ & FDFR$\uparrow$ & ISM$\downarrow$ & SER-FQA$\downarrow$ & BRISQUE$\uparrow$ \\
         \hline
         0 &0.07 &0.63 &0.73 &15.61 &0.21 &0.48 &0.71 &9.64 \\
         0.01 &0.09 &0.58 &0.73 &31.58 &0.28 &0.46 &0.71 &15.85 \\
         0.03 &0.45 &0.39 &0.38 & \textbf{37.82} &0.53 &0.33 &0.47 &38.17 \\
         $0.05^*$ &0.56 &0.33 &0.31 &36.61 &0.62 &0.29 &0.37 &38.22  \\
         0.10 &0.70 &0.22 &0.23 &36.60 &0.77 &0.27 &0.29 &38.59 \\
         0.15 & \textbf{0.77}  &\textbf{0.20} & \textbf{0.20} &36.16 &\textbf{0.83} & \textbf{0.22} & \textbf{0.26} & \textbf{39.17} \\
         \toprule
         \multicolumn{9}{c}{\textbf{CelebA-HQ}} \\
         \hline
         \multirow{2}{*}{$\eta$} & \multicolumn{4}{|c|}{``a photo of \textit{sks} person''} & \multicolumn{4}{c}{``a dslr portrait of \textit{sks} person''}\\
         \cline{2-9}
          & FDFR$\uparrow$ & ISM$\downarrow$ & SER-FQA$\downarrow$ & BRISQUE$\uparrow$ & FDFR$\uparrow$ & ISM$\downarrow$ & SER-FQA$\downarrow$ & BRISQUE$\uparrow$ \\
         \hline
         0 &0.10 &0.68 &0.72 &17.06 &0.26 &0.44 &0.72 &7.30 \\
         0.01 &0.12 &0.68 &0.73 &19.55 &0.30 &0.46 &0.71 &6.60 \\
         0.03 &0.15 &0.57 &0.71 &33.89 &0.27 &0.41 &0.73 &22.67 \\
         $0.05^*$ &0.34 &0.48 &0.56 &36.13 &0.35 &0.36 &0.66 &33.60 \\
         0.10 &0.73 &0.32 &0.27 & \textbf{39.16} &0.67 &0.24 &0.43 & \textbf{38.99} \\
         0.15 & \textbf{0.77} & \textbf{0.29} & \textbf{0.26} &38.22 &\textbf{0.73} & \textbf{0.23} & \textbf{0.35} &38.22 \\
         \bottomrule
    \end{tabular}
    \vspace{-3mm}
    \caption{Defense performance of FSMG with different noise budgets on VGGFace2 and CelebA-HQ in a convenient setting. ``*'' is default.}
    \vspace{-3mm}
    \label{tab:fsmg_noise}
\end{table*}

\begin{table*}[t]
    \centering
    \setlength{\tabcolsep}{3pt}
    \begin{tabular}{c|cccc|cccc}
        \hline
         \multirow{2}{*}{$\eta$} & \multicolumn{4}{|c|}{``a photo of \textit{sks} person''} & \multicolumn{4}{c}{``a dslr portrait of \textit{sks} person''}\\
         \cline{2-9}
          & FDFR$\uparrow$ & ISM$\downarrow$ & SER-FQA$\downarrow$ & BRISQUE$\uparrow$ & FDFR$\uparrow$ & ISM$\downarrow$ & SER-FQA$\downarrow$ & BRISQUE$\uparrow$ \\
         \hline
         0 &0.10 &0.68 &0.72 &17.06 &0.26 &0.44 &0.72 &7.30 \\
         0.01 &0.11 &0.67 &0.72 &19.97 &0.27 &0.45 &0.72 &6.65 \\
         0.03 &0.12 &0.60 &0.71 &34.34 &0.25 &0.44 &0.73 &18.29 \\
         $0.05^*$ &0.31 &0.50 &0.55 &38.57 &0.34 &0.39 &0.63 &34.89  \\
         0.10 &0.73 &0.36 &0.30 &\textbf{38.83} &0.74 &0.27 &0.36 &\textbf{38.96} \\
         0.15 &\textbf{0.86} & \textbf{0.25} & \textbf{0.19} &38.67 & \textbf{0.82} & \textbf{0.24} & \textbf{0.28} &38.86 \\
         \hline
    \end{tabular}
    \vspace{-3mm}
    \caption{Defense performance of ASPL with different noise budgets on CelebA-HQ in a convenient setting. ``*'' is default.}
    \vspace{-3mm}
    \label{tab:aspl_noise}
\end{table*}

\begin{table*}[t]
    \centering
    \setlength{\tabcolsep}{2.5pt}
    \begin{tabular}{l|c|c|cccc|cccc}
        \hline
          & \multirow{2}{*}{Train} & \multirow{2}{*}{Test} & \multicolumn{4}{|c|}{``a photo of \textit{sks} person''} & \multicolumn{4}{c}{``a dslr portrait of \textit{sks} person''}\\
         \cline{4-11}
          & & & FDFR$\uparrow$ & ISM$\downarrow$ & SER-FQA$\downarrow$ & BRISQUE$\uparrow$ & FDFR$\uparrow$ & ISM$\downarrow$ & SER-FQA$\downarrow$ & BRISQUE$\uparrow$ \\
         \hline
         \multirow{2}{.076\linewidth}{Model mismatch} & v1.4 & v2.1 &0.37 &0.48 &0.53 &39.28 &0.34 &0.39 &0.64 &33.50 \\
         & v1.4, 1.5, 2.1 & v2.1 &\textbf{0.39} & \textbf{0.46} &\textbf{0.48} &\textbf{38.25} &\textbf{0.44} &\textbf{0.34} &\textbf{0.57} &\textbf{37.29} \\
         \hline
         \multirow{2}{.076\linewidth}{Ensemble} & v1.4 & v2.0 &0.40 &0.46 &0.51 &38.88 &0.43 &0.36 &0.60 &22.21 \\
         & v1.4, 1.5, 2.1 & v2.0 & \textbf{0.56} & \textbf{0.43} & \textbf{0.43} &\textbf{41.83} &\textbf{0.55} &\textbf{0.33} &\textbf{0.51} &\textbf{29.93} \\
         \hhline{===========}
         \multirow{4}{.076\linewidth}{Term/ Prompt mismatch} & \multicolumn{2}{|c|}{\multirow{2}{*}{DreamBooth prompt}} & \multicolumn{4}{|c|}{``a photo of \textit{$S_*$} person''} & \multicolumn{4}{c}{``a dslr portrait of \textit{$S_*$} person''}\\
         \cline{4-11}
          & \multicolumn{2}{|c|}{} &FDFR$\uparrow$ & ISM$\downarrow$ & SER-FQA$\downarrow$ & BRISQUE$\uparrow$ & FDFR$\uparrow$ & ISM$\downarrow$ & SER-FQA$\downarrow$ & BRISQUE$\uparrow$ \\
         \cline{2-11}
         & \multicolumn{2}{|c|}{``\textit{sks}'' $\rightarrow$ ``\textit{t@t}''} &0.20 &0.17 &0.64 &26.49 &0.17 &0.10 &0.65  &1.14 \\
         & \multicolumn{2}{|c|}{``a dslr portrait of \textit{sks} person''} &0.13 &0.22 &0.69 &18.51 &0.33 &0.51 &0.58 &37.99 \\
         \hline
    \end{tabular}
    \vspace{-3mm}
    \caption{Defense performance of ASPL on CelebA-HQ when the model, term, or prompt used to train the target DreamBooth model is different from the one used to generate defense noise and when the ensemble technique is applied. Here, \textit{$S_*$} is ``t@t'' for the first row and ``sks'' for second row.}
    \label{tab:aspl_cross}
    \vspace{-1mm}
\end{table*}

\begin{table*}[t]
    \centering
    \setlength{\tabcolsep}{3pt}
    \begin{tabular}{c|c|cccc|cccc}
        \hline
         \multirow{2}{*}{Perturbed} & \multirow{2}{*}{Clean} & \multicolumn{4}{|c|}{``a photo of \textit{sks} person''} & \multicolumn{4}{c}{``a dslr portrait of \textit{sks} person''}\\
         \cline{3-10}
          & & FDFR$\uparrow$ & ISM$\downarrow$ & SER-FQA$\downarrow$ & BRISQUE$\uparrow$ & FDFR$\uparrow$ & ISM$\downarrow$ & SER-FQA$\downarrow$ & BRISQUE$\uparrow$ \\
         \hline
         4&0 &\textbf{0.31} &\textbf{0.50} & \textbf{0.55} &\textbf{38.57} &\textbf{0.34} & \textbf{0.39} & \textbf{0.63} & \textbf{34.89} \\
         \hline
         3&1 &0.26 &0.54 &0.63 &32.23 &0.30 &0.40 &0.69 &22.03 \\
         2&2 &0.19 &0.61 &0.69 &25.14 &0.25 &0.41 &0.71 &11.35 \\
         1&3 &0.13 &0.65 &0.72 &19.24 &0.23 &0.43 &0.72 &9.70 \\
         \hline
         0&4 &0.10 &0.68 &0.72 &17.06 &0.26 &0.44 &0.72 &7.30 \\
         \hline
    \end{tabular}
    \vspace{-3mm}
    \caption{Defense performance of ASPL on CelebA-HQ in uncontrolled settings. We include two extra results with 0 clean image (convenient setting) and 0 perturbed image (no defense) for comparison.} 
    \vspace{-3mm}
    \label{tab:aspl_uncontrolled}
\end{table*}

\end{document}